%% file: main.tex
\documentclass[11pt, a4paper, logo, copyright]{googledeepmind}

\usepackage[authoryear, sort&compress, round]{natbib}

\usepackage{microtype}
\usepackage{graphicx}
\usepackage{booktabs} 

\usepackage{amsmath}
\usepackage{amssymb}
\usepackage{mathtools}
\usepackage{amsthm}
\usepackage{caption}
\usepackage{subcaption}

\usepackage[capitalize,noabbrev]{cleveref}

\theoremstyle{plain}

\theoremstyle{definition}

\theoremstyle{remark}

\DeclareMathOperator*{\argmin}{argmin}

\DeclareMathOperator*{\Mtrue}{\mathcal{M}_{\text{env}}}
\newcommand{\model}{\mathcal{M}}

\newcommand{\eat}[1]{} 
\newcommand{\btf}{block teacher forcing}
\newcommand{\btfac}{BTF}

\RequirePackage{times}

\RequirePackage{fancyhdr}
\RequirePackage{xcolor} 
\RequirePackage{algorithm}
\RequirePackage{algorithmic}
\RequirePackage{natbib}
\RequirePackage{eso-pic} 
\RequirePackage{forloop}
\RequirePackage{url}

\title{Improving Transformer World Models for Data-Efficient RL}

\correspondingauthor{adedieu@google.com, joeortiz@google.com}

\renewcommand{\today}{2025-02-01}

\author[*,1]{Antoine Dedieu}
\author[*,1]{Joseph Ortiz}
\author[1]{Xinghua Lou}
\author[1]{Carter Wendelken}
\author[1]{Wolfgang Lehrach}
\author[1]{J. Swaroop Guntupalli}
\author[1]{Miguel Lazaro-Gredilla}
\author[1]{Kevin Murphy}

\affil[*]{Equal contributions}
\affil[1]{Google DeepMind}

\input{arxiv/abstract-v5}

\begin{document}

\maketitle

\input{arxiv/intro-v3}

\input{arxiv/related-v3}
\input{arxiv/methods-v2}
\input{arxiv/expts-v2}

\input{arxiv/concl}

\bibliographystyle{abbrvnat}
\bibliography{arxiv/references}

\input{arxiv/appendix}

\end{document}

%% file: arxiv/abstract-v5.tex
\begin{abstract}
We present three improvements to the standard model-based RL paradigm based on transformers:
(a) ``Dyna with warmup'', which trains the policy on real and imaginary data, but only starts using imaginary data after the world model has been sufficiently trained;
(b) ``nearest neighbor tokenizer'' for image patches, which improves upon previous tokenization schemes, which are
needed when using a transformer world model (TWM),
by ensuring the code words are static after creation, thus providing a constant target for TWM learning; 
and
(c) ``block teacher forcing'', which allows the TWM to reason jointly about the future tokens of the next timestep, instead of generating them sequentially.

We then show that our method significantly improves upon prior methods in various environments.
We mostly focus on  the challenging Craftax-classic benchmark,
where our method achieves a reward of $69.66\%$ after only $1$M environment steps, significantly outperforming DreamerV3, which achieves $53.2\%$,
and exceeding human performance of $65.0\%$ for the first time. 
We also show preliminary results on Craftax-full, MinAtar,
and three different two-player games, to illustrate the generality of the approach.
\end{abstract}

%% file: arxiv/intro-v3.tex
\section{Introduction}
\label{sec:introduction}

Reinforcement learning (RL) \citep{sutton2018reinforcement} 
provides a framework for training agents to act in environments so as to maximize their rewards. Online RL algorithms interleave taking actions in the environment---collecting observations and rewards---and updating the policy using the collected experience. 
Online RL algorithms often employ a model-free approach (MFRL), where the agent learns a direct mapping from observations to actions,
but this can require a lot of data to be collected from the environment.
Model-based RL (MBRL) aims to reduce the amount of data needed to train the policy
 by also learning a world model (WM), and using this WM to plan ``in imagination".

\begin{figure}[t!]
    \centering
    \begin{tabular}{c}
    \includegraphics[width=.5\linewidth]{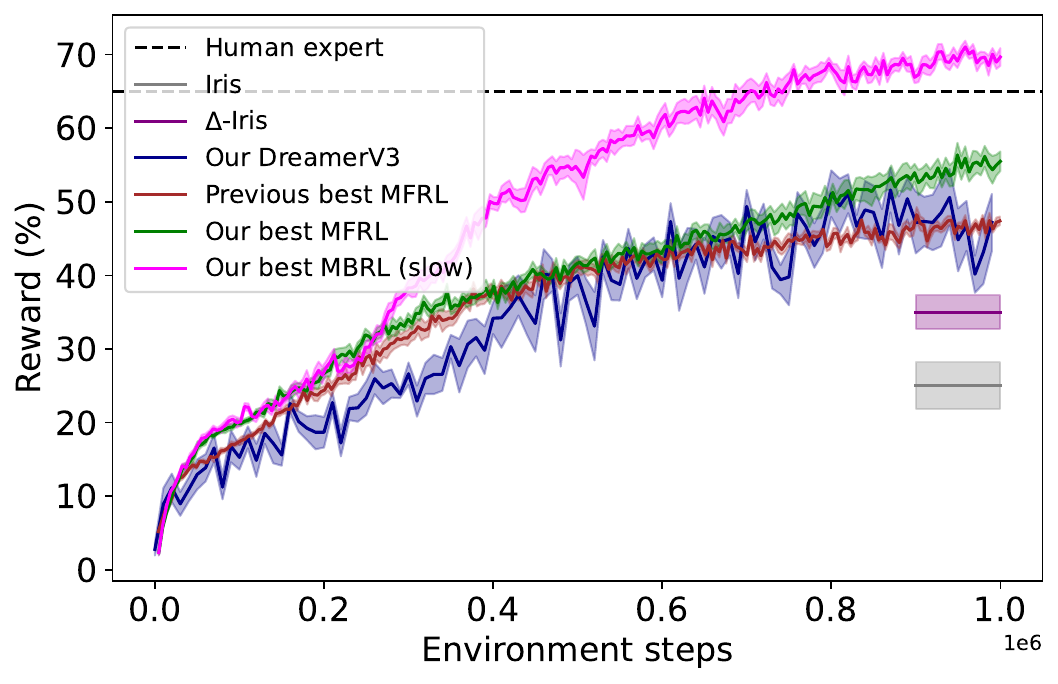}
    \end{tabular}
    \hspace{-.5em}
    \begin{tabular}{c}
        \includegraphics[width=.2\linewidth]{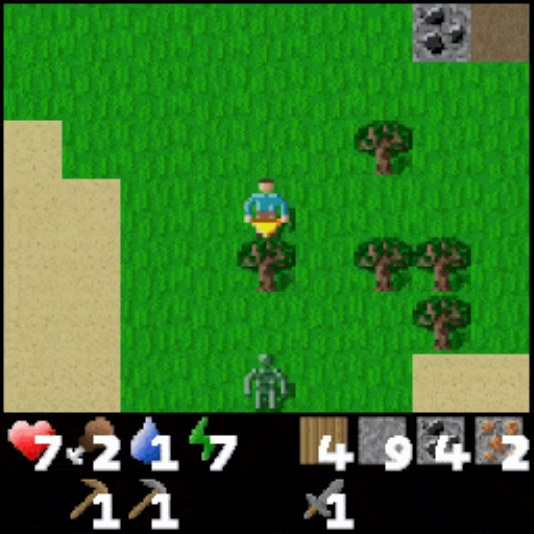} 
    \end{tabular}
    \hspace{-1em}
    \begin{tabular}{c}
        \includegraphics[width=.21\linewidth]{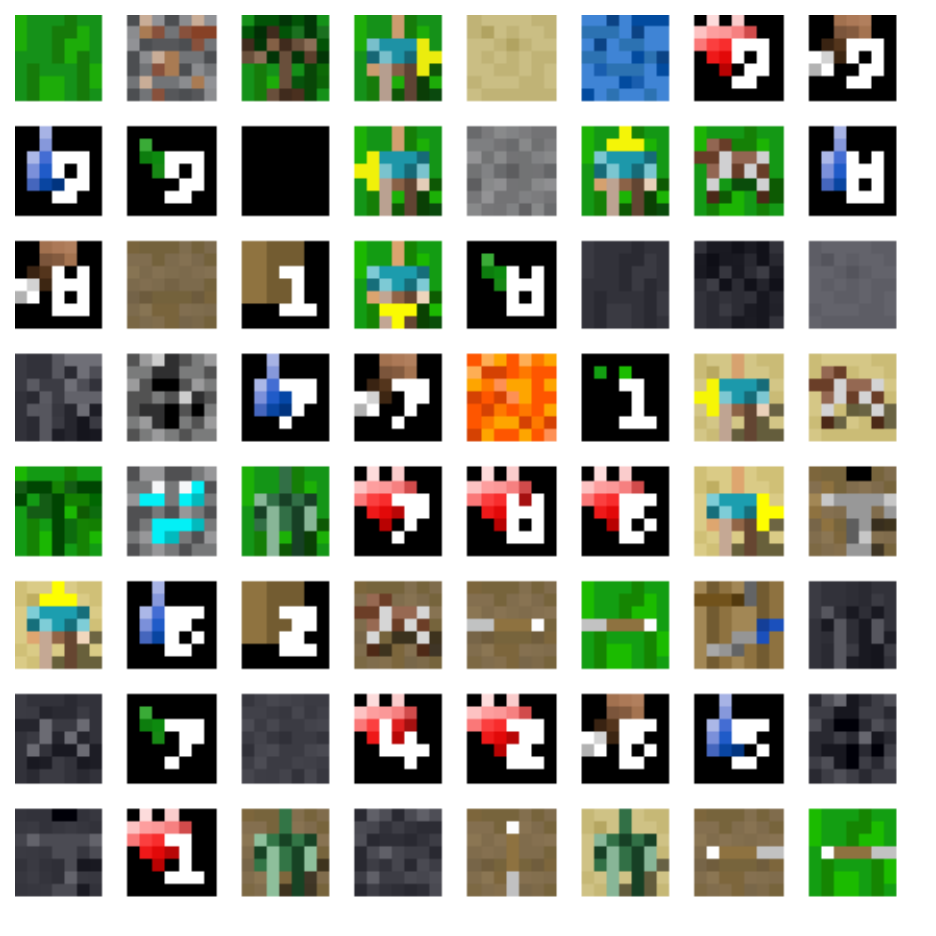}
    \end{tabular}
    \vspace{-.75em}
    \caption{
    [Left]
    Reward on Craftax-classic.
    Our best MBRL and MFRL agents outperform all the previously published MFRL and MBRL results, and for the first time, surpass the reward achieved by a human expert.
    We display published methods which report the reward at 1M steps with horizontal line from 900k to 1M steps. 
    [Middle] The Craftax-classic observation is a $63 \times 63$ pixel image, composed of $9\times9$ patches of $7\times7$ pixels. 
    The observation shows the map around the agent and the agent's health and inventory. Here we have rendered the image at $144 \times 144$ pixels for visibility. 
    [Right] $64$ different patches.
    }
    \label{fig:teaser}
\vspace{-1.em}
\end{figure}

To evaluate sample-efficient RL algorithms, it is common to use the
Atari-$100$k benchmark \citep{Kaiser2019}. 
However, although the benchmark encompasses a variety of  skills (memory, planning, etc), each individual game typically only emphasizes one or two such skills.
To promote the development of agents with broader capabilities, we focus on the Crafter domain \citep{hafner2021benchmarking},
a 2D version of Minecraft that challenges a single agent to master a diverse skill set.
Specifically, we use the  Craftax-classic environment \citep{matthews2024craftax},  a fast, near-replica of Crafter, implemented in JAX \citep{jax2018github}.
Key features of Craftax-classic include: 
(a) procedurally generated stochastic environments (at each episode the agent encounters a new environment sampled from a common distribution); 
(b) partial observability, as the agent only sees a $63 \times 63$ pixel image representing a local view of the agent's environment, plus a visualization of its inventory (see \cref{fig:teaser}[middle]);
and (c) an achievement hierarchy that defines a sparse reward signal, requiring deep and broad exploration.

In this paper, we study improvements to MBRL methods,
based on transformer world models (TWM),
in the context of the Craftax-classic environment.
\eat{
In particular, we address three main questions:
(1) What is the form of the world model (WM)?;
(2) How is the WM trained?;
(3) How is the WM used?.
This paper makes a contribution to each one of these questions.
}
We make contributions across
the following three axes:
(a) how the TWM is used (Section \ref{sec:dyna});
(b) the tokenization scheme used to create TWM inputs
(Section \ref{sec:nnt});
(c) and how the TWM is trained (Section \ref{sec:btf}).
Collectively, our improvements result in an agent that,
with only $1$M environment steps,
achieves a Craftax-classic reward of $69.66\%$ and a score of $31.77\%$,
significantly improving over the previous state of the art (SOTA) reward of $53.20\%$ \citep{hafner2023mastering} and the previous SOTA score of $19.4\%$ \citep{Kauvar2023}\footnote{The score $S$ is given by the geometric mean
of the success rate $s_i$ for each of the
$N=22$ achievements;
this
 puts more weight on occasionally
solving many achievements than on consistently solving
a subset.
More precisely, the score is given by
$S = \exp \left(\frac{1}{N} \sum_{i=1}^N
\ln (1+s_i) \right)-1$,
where $s_i \in [0,100]$ is the success percentage
for achievement $i$
(i.e., fraction of episodes in which
the achievement was obtained at least once).
By contrast, the rewards are just the expected sum of rewards, or in percentage, the arithmetic mean
$R=\frac{1}{N} \sum_{i=1}^N s_i$
(ignoring minor contributions to
the reward based on
 the health of the agent).
 The score and reward are correlated, but are not the same. Unlike some prior work, we report both metrics to make comparisons easier.
}.

\eat{
Our first contribution is to the form 
of the world model.
Following prior work, we 
 use transformers
\citep{vaswani2017attention}
to create a Transformer World Model (TWM).
However, 
since our goal is training agents with visual input,
we need to address the issue of
where the tokens come from.
}

Our first contribution relates to the way the world model is used:
in contrast to recent MBRL methods like IRIS \citep{micheli2022transformers} and DreamerV3 \citep{hafner2023mastering}, which train the policy solely on imagined trajectories (generated by the world model), we train our policy using both imagined rollouts from the world model and real experiences collected in the environment.
This is similar to the original Dyna method \citep{sutton1990integrated}, although this technique has been abandoned in recent work.
In this hybrid regime, we can view the WM as a form
of generative data augmentation
\citep{van2019use}.

Our second contribution addresses the tokenizer which converts between images and tokens that the TWM ingests and outputs.
Most prior work uses a vector quantized variational autoencoder (VQ-VAE, \citealt{van2017neural}), e.g.
IRIS \citep{micheli2022transformers},
DART \citep{agarwal2024learning}.
\eat{
These methods train a CNN to map images $O_t$ to a 
feature map $Z_t$, whose elements $Z_{t}^i$ are then quantized into a set of discrete tokens
$Q_t=\{q_{t}^i\}_{i=1}^L$, where $q_{t}^{i} \in \{1,\ldots,K\}$
is the index of the codebook vector representing
$Z_{t}^i$,
and $i \in \{1,\ldots,L\}$ is the token index.
The tokens $Q_t$ are concatenated into a sequence $Q_{1:T}$ to represent the observations
which, along with the sequence of actions, is used to train the WM.
}
These methods train a CNN to process images into a 
feature map, whose elements are then quantized into discrete tokens, using a codebook. 
The sequence of observation tokens across timesteps is used, along with the actions and rewards, to train the WM.
We propose two improvements to the tokenizer.
First, instead of jointly quantizing the image, we split the image into patches and independently tokenize each patch.
Second, we replace the VQ-VAE with a simpler nearest-neighbor tokenizer (NNT) for patches. 
Unlike VQ-VAE, NNT ensures that the ``meaning" of each code in the codebook is constant through training, which simplifies the task of learning a reliable WM.

Our third contribution addresses the way the world model is trained.
TWMs are trained by maximizing the log likelihood of the sequence of tokens,
which is typically generated autoregressively both over time and within a timeslice.
We propose an alternative, 
which we call \btf ~(BTF),
that allows TWM to reason jointly about the possible future states of all tokens within a timestep, before sampling them in parallel and independently given the history.
With BTF, imagined rollouts for training the policy are both faster to sample and more accurate.

Our final contributions are some minor architectural changes to the MFRL baseline upon which our MBRL
approach is based. These changes are still significant, resulting in a simple MFRL method that is much faster than Dreamer V3 and yet obtains a much better average reward and score.

Our improvements are complementary to each other, and  can be combined into a  ``ladder of improvements"---similar to the ``Rainbow" paper's
\citep{Hessel2018} series of improvements on top of model-free DQN agents. 


\eat{
It is also consistent
with the results in \citep{Jesson2024},
which  shows that minor architectural
changes to the policy,
trained with PPO without a WM,
can lead to big gains on the ProcGen benchmark.
} %

%% file: arxiv/related-v3.tex
\section{Related Work}
\label{sec:related}

In this section, we discuss related work in MBRL --- see e.g. \citet{Moerland2023,murphy2024reinforcement, awesome_mbrl} 
for more comprehensive reviews.
We can broadly divide MBRL along two axes.
The first axis is whether the world model (WM) is used for background planning
(where it helps train the policy by generating imagined trajectories),
or decision-time planning (where it is used for lookahead search at inference time). 
The second axis is whether the WM is a generative model of the observation space (potentially via a latent bottleneck) or whether is a latent-only model trained using a self-prediction loss
(which is not sufficient to generate full observations).

Regarding the first axis, prominent examples of decision-time planning methods that leverage a WM include MuZero \citep{schrittwieser2020mastering} and EfficientZero  \citep{Ye2021}, which use Monte-Carlo tree search over a discrete action space, as well as TD-MPC2 \citep{Hansen2024}, which uses the cross-entropy method over a continuous action space.
Although some studies have shown that decision-time planning can sometimes be better
than background planning \citep{Alver2024}, it is much slower, especially with large WMs such as transformers, since it requires rolling out future hypothetical trajectories at each decision-making step.
Therefore in this paper, we focus on background planning (BP).
Background planning originates from Dyna \citep{sutton1990integrated}, which focused on tabular Q-learning.
Since then, many papers have combined the idea with deep RL methods: World Models \citep{ha2018recurrent},
Dreamer agents  \citep{Hafner2020, hafner2020mastering,hafner2023mastering},
SimPLe \citep{Kaiser2019},
IRIS \citep{micheli2022transformers},
$\Delta$-IRIS \citep{micheli2024efficient},
Diamond \citep{alonso2024diffusion},
DART \citep{agarwal2024learning}, etc.

Regarding the second axis, 
many methods fit generative WMs of the
observations (images) using a 
model with low-dimensional latent variables,
either continuous (as in a VAE)
or discrete (as in a VQ-VAE).
This includes our method and most background planning methods above
\footnote{
A notable exception is 
Diamond \citep{alonso2024diffusion},
which fits a diffusion world model
directly in pixel space,
rather than learning a latent WM. 
}.
In contrast, other methods fit non-generative WMs, which are trained using self-prediction loss---see \citet{Ni2024} for a detailed discussion.
Non-generative WMs are more lightweight and therefore well-suited to decision-time planning with its large number of WM calls at every decision-making step.
However,
generative WMs are generally preferred for background planning, since it is easy to combine real and imaginary data for policy learning,
as we show below.

In terms of the WM architecture, many state-of-the-art models use transformers, e.g. 
IRIS \citep{micheli2022transformers},
$\Delta$-IRIS \citep{micheli2024efficient},
DART \citep{agarwal2024learning}.
Notable exceptions are DreamerV2/3  \citep{hafner2020mastering,hafner2023mastering}, which use recurrent state space models,
although improved transformer variants have been proposed 
\citep{robine2023transformer,zhang2024storm,chen2202transdreamer}.

%% file: arxiv/methods-v2.tex
\section{Methods}
\label{sec:methods}


\subsection{MFRL Baseline}
\label{sec:MFRL}

Our starting point is the previous SOTA
MFRL approach which was proposed as a baseline in
\citet{moon2024discovering}\footnote{%
The authors' main method uses external knowledge about the achievement hierarchy of Crafter, so cannot be compared with other general methods. We use their baseline instead.
}.
This method achieves a reward of $46.91\%$ and a score of $15.60\%$ after $1$M environment steps.
This approach trains a stateless CNN policy without frame stacking
using the PPO method \citep{schulman2017proximal}, and
adds an entropy penalty to ensure sufficient exploration.
The CNN used is a modification of the Impala
ResNet \citep{Espeholt2018}.

\subsection{MFRL Improvements}

We improve on this MFRL baseline
by both increasing the model size and adding a RNN (specifically a GRU) to give the policy memory.
Interestingly, we find that naively increasing the model size harms performance, while combining a larger model with a carefully designed RNN helps (see Section \ref{sec:ablations}).
When varying the ratio of the RNN state dimension to the CNN encoder dimension, we observe that performance is higher when the hidden state is low-dimensional. Our intuition is that the memory is forced to focus on the relevant bits of the past that cannot be extracted from the current image.
We concatenate the GRU output to the image
embedding, and then pass this to the actor and critic networks,
rather than directly passing the GRU output. Algorithm \ref{algo:ac_network}, Appendix \ref{ap:mfrl_agent}, presents a pseudocode for our MFRL agent.

With these architectural changes,
we increase the reward to $55.49\%$ and the score to $16.77\%$.
This result is notable since our MFRL agent beats the considerably more complex (and much slower) DreamerV3 agent, which obtains a reward of $53.20\%$ and a score of $14.5$. 
It also beats other MBRL methods, such as IRIS \citep{micheli2022transformers} (reward of $25.0\%$) and $\Delta$-IRIS \citep{micheli2024efficient}
\footnote{
This is consistent with results on Atari-100k, which show that well-tuned model-free methods, such as BBF \citep{Schwarzer2023}, can beat more sophisticated model-based methods.} (reward of $35.0\%$). 
In addition, our MFRL agent only takes
$15$ minutes to train for $1$M environment
steps on one A100 GPU.


\subsection{MBRL baseline}
\label{sec:IRIS}
\label{sec:MBRLbaseline}

We now describe our MBRL baseline,
which combines our MFRL baseline above with a transformer world model (TWM)---as in
IRIS \citep{micheli2022transformers}. Following IRIS, our MBRL baseline uses a VQ-VAE, which quantizes the
$8 \times 8$ feature map $Z_t$ of a CNN
to create a set of latent codes,
$(q_t^1,\ldots,q_t^L) = \text{enc}(O_t)$,
where $L=64$, $q_t^i \in \{1,\ldots,K\}$
is a discrete code,
and $K=512$ is the size of the codebook.
These codes are then passed to a TWM, which is trained using
teacher forcing---see \cref{eqn:lossAR} below. 
Our MBRL baseline achieves a reward of $31.93\%$, and improves over the reported results of IRIS, which reaches $25.0\%$. 

Although these MBRL baselines leverage recent advances in generative world modeling, they are largely outperformed by our best MFRL agent. This motivates us to enhance our MBRL agent, which we explore in the following sections.


\subsection{MBRL using Dyna with warmup}
\label{sec:dyna}
\label{sec:warmup}

As discussed in \cref{sec:introduction}, we propose to train our MBRL agent on a mix of real trajectories (from the environment) and imaginary trajectories (from the TWM), similar to Dyna \citep{sutton1990integrated}.
\cref{algo:MBRL} presents the pseudocode for our MBRL approach.
Specifically, unlike many other recent MBRL methods \citep{Ha2018, micheli2022transformers, micheli2024efficient, hafner2020mastering, hafner2023mastering} which train their policies exclusively using world model rollouts (Step 4), we include Step 2 which updates the policy with real trajectories.
Note that, if we remove Steps 3 and 4 in Algorithm \ref{algo:MBRL}, the approach reduces to MFRL.
The function $\text{rollout}(O_1, \pi_{\Phi},  T, \model)$ returns a trajectory of length $T$ generated by rolling out the policy $\pi_{\Phi}$ from the initial state $O_1$ in either the true environment $\Mtrue$ or the world model $\mathcal{M}_{\Theta}$.
A trajectory contains collected observations, actions and rewards during the rollout $\tau = (O_{1:T+1}, a_{1:T}, r_{1:T})$.
Algorithm \ref{algo:rollout} in Appendix \ref{ap:mbrl_agent} details the rollout procedure.
We discuss other design choices below. 

\begin{algorithm}[!htbp]
\caption{MBRL agent. See Appendix \ref{ap:mbrl_agent} for details.}
\label{algo:MBRL}
\begin{algorithmic}
\STATE {\bfseries Input:} number of environments $N_{\text{env}}$,
\newline
environment dynamics $\Mtrue$,
\newline
rollout horizon for environment $T_{\text{env}}$ and for TWM $T_{\text{WM}}$,
\newline
background planning starting step $T_{\text{BP}}$,
\newline
total number of environment steps $T_{\text{total}}$,
\newline
number of TWM updates $N^{\text{iters}}_{\text{WM}}$ and policy updates $N^{\text{iters}}_{\text{AC}}$
\medskip
\STATE {\bfseries Initialize:} observations $O^n_1 \sim \Mtrue ~\text{for}~ \small{n=1:N_{\text{env}}}$,
\newline
data buffer $\mathcal{D}=\emptyset$, 
\newline
TWM model $\mathcal{M}$ and parameters $\Theta$, 
\newline
AC model $\pi$ and parameters $\Phi$,
\newline
number of environment steps $t=0$.
\smallskip
\REPEAT
\STATE \textit{// 1. Collect data from environment}
\STATE $\tau^n_{\text{env}} = \text{rollout}
(O_1^n, \pi_{\Phi}, T_{\text{env}}, \Mtrue),~~n=1:N_{\text{env}}$
\STATE $\mathcal{D} = \mathcal{D} \cup \tau^{1:N}_{\text{env}} ~;~ O^{1:N}_1 = \tau^{1:N}_{\text{env}}[-1] ~;~ t += N_{\text{env}}T_{\text{env}}$
\medskip
\STATE \textit{// 2. Update policy on environment data}
\STATE $\Phi=\text{PPO-update-policy}(\Phi,\tau_{\text{env}}^{1:N})$\\
\medskip
\STATE \textit{// 3. Update world model}
\FOR{$\text{it}=1$ {\bfseries to} $N^{\text{iters}}_{\text{WM}}$}
\STATE  $\tau_{\text{replay}} ^{n} =\text{sample-trajectory}(\mathcal{D}, T_{\text{WM}}), ~~n=1:N_{\text{env}}$ \\
\STATE  $\Theta = \text{update-world-model}(\Theta, \tau_{\text{replay}}^{1:N_{\text{env}}})$\\
\ENDFOR
\medskip
\STATE \textit{// 4. Update policy on imagined data}
\IF{ $t \ge T_{\text{BP}}$}
\FOR{$\text{it}=1$ {\bfseries to} $N^{\text{iters}}_{\text{AC}}$}
\STATE $\tilde{O}_1^n = \text{sample-obs}(\mathcal{D}), ~~n=1:N_{\text{env}}$ 
\STATE $\small{\tau_{\text{WM}}^{n}=\text{rollout}
(\tilde{O}_1^n,\pi_{\Phi}, T_{\text{WM}},\mathcal{M}_{\Theta}), ~n=1:N_{\text{env}}}$ 
\STATE $\Phi=\text{PPO-update-policy}(\Phi,\tau_{\text{WM}}^{1:N_{\text{env}}})$
\ENDFOR
\ENDIF
\UNTIL $t \ge T_{\text{total}}$
\end{algorithmic}
\end{algorithm}

\textbf{PPO.}
Since PPO \citep{schulman2017proximal} is an on-policy algorithm, trajectories should be used for policy updates immediately after they are collected or generated. 
For this reason, policy updates with real trajectories take place in Step 2 immediately after the data is collected. 
An alternative approach is to use an off-policy algorithm and mix real and imaginary data into the policy updates in Step 4, hence removing Step 2. We leave this direction as future work. 

\textbf{Rollout horizon.}
We set $T_{\text{WM}} \ll  T_{\text{env}}$,
to avoid the problem of compounding errors
due to model imperfections \citep{Lambert2022}.
However, we find it beneficial to use
$T_{\text{WM}} \gg 1$,
consistent with
\citet{Holland2018,van2019use},
who observed that the
Dyna approach with $T_{\text{WM}}=1$
is no better than
MFRL with experience replay.

\textbf{Multiple updates.} Following IRIS, we update TWM $N^{\text{iters}}_{\text{WM}}$ times and the policy on imagined trajectories $N^{\text{iters}}_{\text{AC}}$ times.

\textbf{Warmup.}
When mixing imaginary trajectories with real ones, we need to ensure the WM is sufficiently accurate so that it does not harm policy learning. 
Consequently, we only begin training the policy
on imaginary trajectories after the agent has interacted with the environment for $T_{\text{BP}}$ steps, which ensures it has seen enough data to learn a reliable WM. We call this technique ``Dyna with warmup''. In \cref{sec:ablations}, we show that removing this warmup, and using $T_{\text{BP}}=0$, drops the reward dramatically, from $67.42\%$ to $33.54\%$. We additionally show that removing the Dyna method (and only training the policy in imagination) drops the reward to $55.02\%$.

\eat{
In \cref{sec:results}, we show the advantage of using Dyna rather than only training the policy
in imagination,
and the advantage of waiting until the WM is warmed up
rather than using it immediately.
}

\subsection{Patch nearest-neighbor tokenizer}
\label{sec:nnt}
\label{sec:patches}

Many MBRL methods based on TWMs use a VQ-VAE to map between images and tokens.
In this section, we describe our alternative which leverages a property of Craftax-classic: each observation is composed of $9\times9$ patches of size $7 \times 7$ each (see \cref{fig:teaser}[middle]). Hence we propose to (a) factorize the tokenizer by patches and (b) use a simpler nearest-neighbor style approach to tokenize the patches.

\paragraph{Patch factorization.}
Unlike prior methods which process the full image $O$ into tokens $(q^1,\ldots,q^L) = \text{enc}(O)$, we first divide $O$ into $L$ non-overlapping patches
$(p^1, \ldots, p^L)$ which are independently encoded into $L$ tokens:
\begin{equation*}
    (q^1,\ldots, q^L) = (\text{enc}(p^1), \ldots, \text{enc}(p^L))
    ~.
\end{equation*}
To convert the discrete tokens back to pixel
space, we just decode each token independently into patches, and rearrange to form a full image:
\begin{equation*}
    (\hat{p}^1, \ldots, \hat{p}^L) = 
    (\text{dec}(q^1), \ldots, \text{dec}(q^L))
    ~.
\end{equation*}
Factorizing the VQ-VAE on the $L=81$ patches of each observation boosts performance from $43.36\%$ to $58.92\%$.

\paragraph{Nearest-neighbor tokenizer.}
On top of patch factorization, we propose a simpler nearest-neighbor tokenizer (NNT) to replace the VQ-VAE.
The encoding operation for each patch $p \in [0, 1]^{h\times w\times3}$
is similar to a nearest neighbor classifier w.r.t the codebook.
The difference is that, if the nearest neighbor
is too far away, 
we add a new code equal to $p$ to the codebook.
More precisely, let us denote  $\mathcal{C}_{\text{NN}}=\{e_1, \ldots,e_K\}$ the current codebook,
consisting of $K$ codes $e_i \in [0, 1]^{h\times w\times3}$, 
and $\tau$ a threshold on the Euclidean distance. 
The NNT encoder is defined as:
\begin{equation}
    \small
    q = \text{enc}(p) = 
    \begin{cases}
    \begin{alignedat}{2}
        &\argmin_{1\le i \le K} \| p -  e_i\|_2^2 &&\quad\text{if $\min_{1\le i \le K} \| p - e_{i} \|_2^2 \le \tau$} \\
        &K+1 &&\quad\text{otherwise.}
    \end{alignedat}
    \end{cases}
    \label{eqn:opt}
\end{equation}
The codebook can be thought of as a
greedy approximation to the coreset of 
the patches seen so far
\citep{Mirzasoleiman2020}. 
To decode patches, we simply return the code associated with the codebook index, i.e. $\text{dec}(q^i)=e_{q^i}$.

A key benefit of NNT is that once codebook entries are added, they are never updated.
A static yet growing codebook makes the target distribution for the TWM stationary, greatly simplifying online learning for the TWM.
In contrast, the VQ-VAE codebook is continually updated, meaning the TWM must learn from a non-stationary distribution, which results in a worse WM.
Indeed, we show in \cref{sec:ladder} that with patch factorization, and when $h=w=7$---meaning that the patches are aligned with the observation---replacing the VQ-VAE with NNT boosts the agent's reward from $58.92\%$ to $64.96\%$. 
Figure \ref{fig:teaser}[right] shows an example of the first 64 code patches extracted by our NNT.

The main disadvantages of our approach are that (a) patch tokenization can be sensitive to the patch size (see Figure \ref{fig:ablation}[left]),
and (b) NNT may create a large codebook if there is a lot
of appearance variation within patches.
In Craftax-classic, these problems are not very severe due to the grid structure of the game and limited sprite vocabulary (although continuous variations exist due to lighting and texture randomness).


\subsection{Block teacher forcing}
\label{sec:btf}

\begin{figure}[h!]
    \centering
    \begin{tabular}{c}
        \includegraphics[width=.45\linewidth]{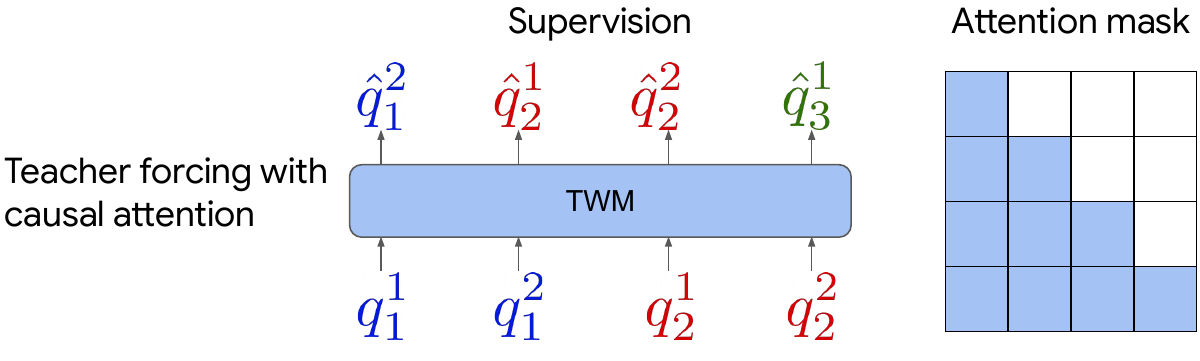} 
    \end{tabular}
    \begin{tabular}{c}
        \includegraphics[width=.47\linewidth]{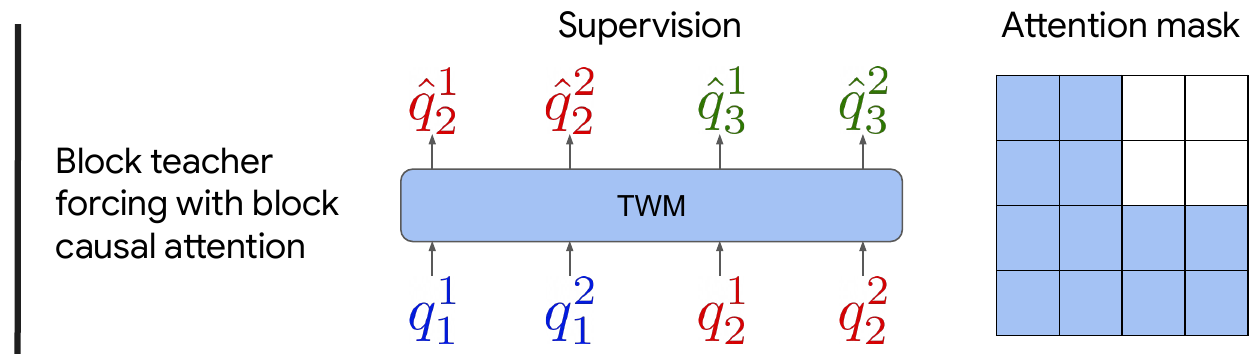}
    \end{tabular}
    \caption{
    Approaches for TWM training with $L=2$, $T=2$.
    $q_t^{\ell}$ denotes token $\ell$ of timestep $t$. Tokens in the same timestep have the same color. 
    We exclude action tokens for simplicity.
    [Left] Usual autoregressive model training with teacher forcing. 
    [Right] Block teacher forcing predicts token $q_{t+1}^{\ell}$ from input token $q_{t}^{\ell}$ with block causal attention.
    }
    \label{fig:btf}
\end{figure}


Transformer WMs are typically trained by teacher forcing which maximizes
the log likelihood of the token sequence generated autoregressively over time
and within a timeslice:
\vspace{-.75em}
\begin{equation}
\small
    \mathcal{L}_{\text{TF}} = 
    \log \prod_{t=1}^{T} \prod_{i=1}^L \mathcal{L}_t^i ~,~~~~
    \mathcal{L}_t^i
    =
    p(q_{t+1}^i | q_{1:t}^{1:L},  q_{t+1}^{1:i-1}, a_{1:t})
    \label{eqn:lossAR}
\end{equation}
We propose a more effective alternative, which we call
\btf~(\btfac).
BTF modifies both the supervision and the attention of the TWM. Given the tokens from the previous timesteps,
BTF independently predicts all the latent tokens at the next timestep, removing the conditioning on previously generated tokens from the current step:
\vspace{-.5em}
\begin{equation}
\small
    \mathcal{L}_{\text{BTF}} = 
    \log \prod_{t=1}^{T} \prod_{i=1}^L \mathcal{\tilde{L}}_t^i ~,~~~~
    \mathcal{\tilde{L}}_t^i
    =
    p(q_{t+1}^i | q_{1:t}^{1:L}, a_{1:t})
    \label{eqn:lossBTF}
\vspace{-.4em}
\end{equation}
Importantly BTF uses a block causal attention pattern (see Figure \ref{fig:btf}), in which tokens within the same timeslice are decoded in-parallel in a single forward pass.
This attention structure allows the model to reason jointly about the possible future states of all tokens within a timestep, before sampling the tokens with independent readouts.
This property mitigates autoregressive drift. As a result, we find that BTF returns more accurate TWMs than fully AR approaches.
Overall, adding BTF increases the reward from $64.96\%$ to $67.42\%$, leading to our best MBRL agent.
In addition, we find that BTF is twice as fast,
even though in theory,
with key-value caching, BTF and AR both have complexity $\mathcal{O}(L^2 T)$ for generating all the $L$ tokens at one timestep, and $\mathcal{O}(L^2 T^2)$ for generating the entire rollout. Finally, BTF shares a similarity with Retentive Environment Models (REMs) \citep{cohen2024improving} in their joint prediction of next-frame tokens. However, while REMs employ a retentive network \citep{sun2023retentive}, BTF offers broader applicability across any transformer architecture.

\eat{
This also seems to slightly
improve the accuracy of the WM and hence the agent's performance: removing \btfac~ drops
the reward from 14.83 to 14.29,
as we show in \cref{sec:results}.
}

%% file: arxiv/expts-v2.tex
\begin{table*}[btp]
\caption{Results on Craftax-classic after 1M environment interactions. 
* denotes results on Crafter,
which may not exactly match Craftax-classic.
--- means unknown.
\textdagger denotes the reported timings on a single A100 GPU.
Our DreamerV3 results are based
on the code from the author,
but differ slightly from the reported
number, perhaps due to 
hyperparameter discrepancies.
IRIS and $\Delta$-IRIS do not report standard errors for the score.
}
\label{tab:best_scores}
\small
\vspace{-.25em}
\centering
    \begin{tabular}{ccccc}
    \toprule
    Method & Parameters & Reward (\%) & Score (\%) & Time (min) \\
    \midrule
    Human Expert & NA & $*65.0 \pm 10.5$ & $*50.5 \pm 6.8$  & NA \\
    \midrule
    M1: Baseline
    & $60.0\text{M}$ & $31.93 \pm 2.22$ & $4.98 \pm 0.50$ & $560$ \\
    M2: M1 + Dyna
    & $60.0$M & $43.36 \pm 1.84$ & $8.85 \pm 0.63$ & $563$ \\
    M3: M2 + patches
    & $56.6$M & $58.92 \pm 1.03$ & $19.36 \pm 1.42$ & $746$ \\
    M4: M3 + NNT
    & $58.5$M & $64.96 \pm 1.13$ & $25.55 \pm 0.86$  & $1328$ \\
    M5: M4 + BTF. Our best MBRL (fast)
    & $58.5$M & $67.42 \pm 0.55$ & $27.91 \pm 0.63$ & $759$ \\
    M5: M4 + BTF. Our best MBRL (slow)
    & $58.5$M & $\mathbf{69.66} \pm 1.20$ & $\mathbf{31.77} \pm 1.43$ & $2749$ \\
    \midrule
    Previous best MFRL \citep{moon2024discovering} 
    & $4.0\text{M}$
    & $*46.91 \pm 2.41$ 
    & $*15.60 \pm 1.66$
    & ---\\
    Previous best MFRL (our implementation) & $4.0\text{M}$ & $47.40 \pm 0.58$ & $10.71 \pm 0.29$ & $26$ \\
    Our best MFRL & $55.6$M & $55.49 \pm 1.33$  & $16.77 \pm 1.11$ & $15$\\
    \midrule
    DreamerV3 \citep{hafner2023mastering} & $201$M  & $*53.2 \pm 8.$ & $*14.5 \pm 1.6$ & --- \\
    Our DreamerV3 & $201$M & $47.18 \pm 3.88$& --- & $2100$ \\
    IRIS \citep{micheli2022transformers} 
    & $48$M & $*25.0 \pm 3.2$  & $*6.66$ & \textdagger$8330$  \\
    $\Delta$-IRIS \citep{micheli2024efficient} & 25M & $*35.0 \pm 3.2$  & $*9.30$  & \textdagger$833$  \\
    Curious Replay \citep{Kauvar2023} & --- & --- & $*19.4 \pm 1.6$ & ---- \\
    \bottomrule
    \end{tabular}
\end{table*}

\section{Results}
\label{sec:results}

In this section, we report our experimental
results on the Craftax-classic benchmark.  Each experiment is run on $8$ H100 GPUs.
All methods are compared after interacting with the environment for
$T_{\text{total}}=1$M steps.
All the methods collect trajectories of length $T_{\text{env}}=96$ in $N_{\text{env}}=48$ environment (in parallel).
For MBRL methods, the imaginary rollouts
are of length $T_{\text{WM}}=20$,
and we start generating these (for policy training) 
after $T_{\text{BP}} = 200\text{k}$ 
environment steps. We update the TWM $N^{\text{iters}}_{\text{WM}}=500$ times and the policy $N^{\text{iters}}_{\text{AC}}=150$ times.
For all metrics, we report the mean and standard error over $10$ seeds as $x(\pm y)$.

\subsection{Climbing up the MBRL ladder}\label{sec:ladder}

\begin{figure}[h!]
    \begin{minipage}{0.48\textwidth} 
    \centering
    \includegraphics[width=.95\linewidth]{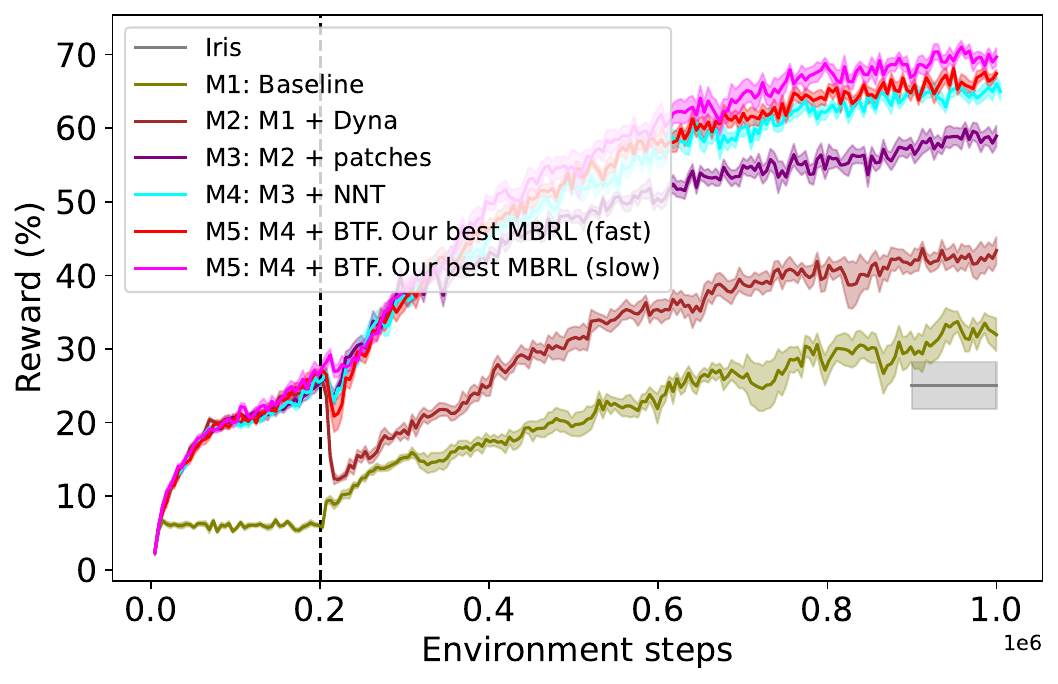}
    \vspace{-0.5em}
    \caption{
    The ladder of improvements presented in Section \ref{sec:methods} progressively transforms our baseline MBRL agent into a state-of-the-art method on Craftax-classic. 
    Training in imagination
    starts at step 200k, indicated
    by the dotted vertical line.
    }
    \label{fig:main}
    \end{minipage}
    \hspace{0.4em}
    \begin{minipage}{0.48\textwidth} 
    \caption{Ablations results on Craftax-classic after 1M environment interactions.}
    \label{tab:ablation}
    \centering
    \resizebox{0.9\textwidth}{!}{
    \begin{tabular}{ccc}
    \toprule
    Method & Reward $(\%)$ & Score $(\%)$ \\
    \midrule
    Our best MBRL (fast)  & $\mathbf{67.42} \pm 0.55$ &
    $\mathbf{27.91} \pm 0.63$ \\
    \midrule
    $5\times5$ quantized  & $57.28 \pm 1.14$ & $18.26 \pm 1.18$ \\
    $9\times9$ quantized  & $45.55 \pm 0.88$ & $10.12 \pm 0.40$ \\
    $7\times7$ continuous & $21.20 \pm 0.55$ & $2.43 \pm 0.09$ \\
    \midrule
    Remove Dyna & $55.02 \pm 5.34$ & $18.79 \pm 2.14$ \\
    Remove NNT  & $60.66 \pm 1.38$ & $21.79 \pm 1.33$ \\
    Remove NNT \& patches & $45.86 \pm 1.42$ & $10.36 \pm 0.69$ \\
    Remove BTF  & $64.96 \pm 1.13$ & $25.55 \pm 0.86$ \\
    \midrule
    Use $T_{\text{BP}}=0$ & $33.54 \pm 10.09$ & $12.86 \pm 4.05$  \\
    \midrule
    \midrule
    Best MFRL  & $55.49 \pm 1.33$ & $16.77\pm 1.11$ \\
    Remove RNN & $41.82 \pm 0.97$ & $8.33 \pm 0.44$ \\
    Smaller model  & $51.35 \pm 0.80$ & $12.93 \pm 0.56$ \\
    \bottomrule
    \end{tabular}
    }
    \end{minipage}
\end{figure}

First, we report the normalized reward (the reward divided by the maximum reward of $22$) for a series of agents that progressively climb our ``MBRL ladder" of improvements in \cref{sec:methods}.
\cref{fig:main} show the reward vs. the number of environment steps for the following methods, which we detail in Appendix \ref{ap:mbrl_modules}:
\smallskip
\newline
$\bullet$ \textbf{M1: Baseline}. Our baseline MBRL agent, described in \cref{sec:MBRLbaseline}, reaches a reward of $31.93\%$, and improves over IRIS, which gets $25.0\%$.
\smallskip
\newline
$\bullet$ \textbf{M2: M1 + Dyna}. Training the policy on both (real) environment and (imagined) TWM trajectories, as described in Section \ref{sec:dyna}, increases the reward to $43.36\%$.
\smallskip
\newline
$\bullet$ \textbf{M3: M2 + patches}. 
Factorizing the VQ-VAE over the $L=81$ observation patches, as presented in Section \ref{sec:nnt}, increases the reward to $58.92\%$.
\smallskip
\newline
$\bullet$ \textbf{M4: M3 + NNT}.
With patch factorization, replacing the VQ-VAE with
NNT, as presented in Section \ref{sec:nnt},
further boosts the reward to $64.96\%$.
\smallskip
\newline
$\bullet$ \textbf{M5: M4 + BTF. Our best MBRL (fast)}: Incorporating BTF, as described in  \cref{sec:btf}, leads to our best agent. It achieves a reward of $67.42\%$, while BTF reduces the training time by a factor of two.
\smallskip
\newline
$\bullet$ \textbf{M5: M4 + BTF. Our best MBRL (slow)}: By increasing the number of TWM training steps to  $N^{\text{iters}}_{\text{WM}}=4$k, we obtain our best agent, which reaches a reward of $69.66\%$. However, due to substantial training times ($\sim 2$ days), we do not include this agent in our ablation studies (Section \ref{sec:ablations}) and comparative studies (Section \ref{sec:stationary}).

As in IRIS \citep{micheli2022transformers}, methods M1-3 use a codebook size of $512$. 
For M4 and M5, which use NNT, we found it critical to use a larger codebook size of $K=4096$ and a threshold of $\tau=0.75$. 
Interestingly, when training in imagination begins (at step $T_{\text{BP}}=200\text{k}$), there is a temporary drop in performance as the TWM rollouts do not initially match the true environment dynamics, resulting in a distribution shift for the policy. 


\subsection{Comparison to existing methods}
\label{sec:cmp}

Figure \ref{fig:teaser}[left]
compares the performance of our best MBRL and MFRL agents against various previous methods. 
See also \cref{fig:score} in Appendix \ref{sec:appendix_score}
for a plot of the score,
and  \cref{tab:best_scores}
for a detailed numerical comparison
of the final performance.
First, we observe that our best MFRL agent outperforms almost
all of the previously published MFRL and MBRL results,
reaching a reward of $55.49\%$
and a score of $16.77\%$\footnote{
The only exception is Curious Replay \citep{Kauvar2023}, which builds on DreamerV3 with prioritized experience replay (PER)
to train the WM.
However, 
PER is only better
on a few achievements;
this improves the score but not the reward.
}.
Second, our best MBRL agent
achieves a new SOTA  reward of $69.66\%$
and a score of $31.77\%$.
This marks the first agent to surpass human-level reward, derived from 100 episodes played by 5 human expert players \citep{hafner2021benchmarking}.
Note that although we achieve superhuman reward, our score is significantly below that of a human expert.

\subsection{Ablation studies}
\label{sec:ablations}

\begin{figure*}[t!]
    \centering
    \begin{tabular}{c}
        \includegraphics[width=0.31\textwidth]{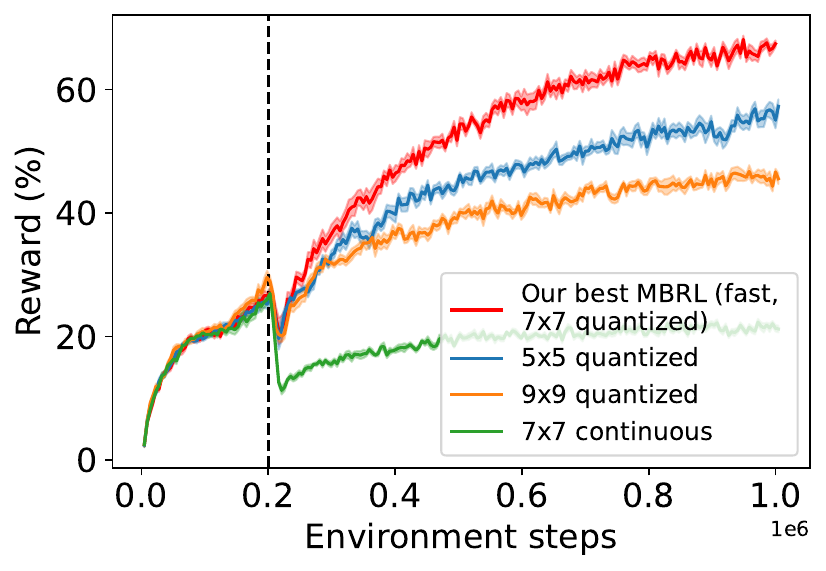}
    \end{tabular}
    \hspace{-5mm}
    \begin{tabular}{c}
        \includegraphics[width=0.31\textwidth]{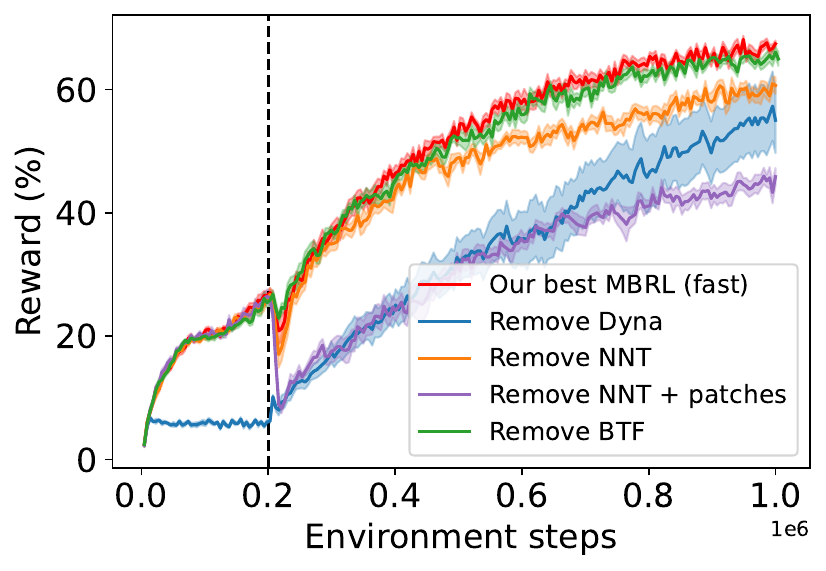}
    \end{tabular}
    \hspace{-5mm}
    \begin{tabular}{c}
        \includegraphics[width=0.31\textwidth]{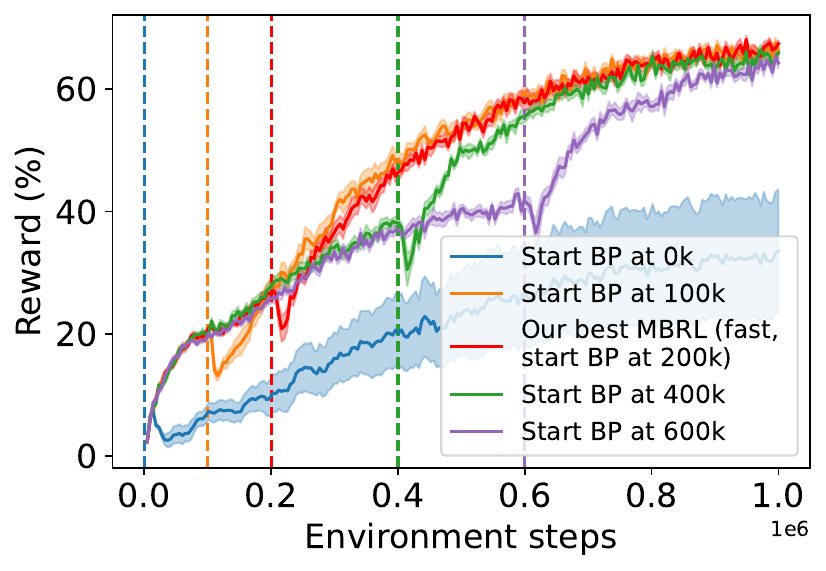}
    \end{tabular}
    \vspace{-.25em}
    \caption{[Left] MBRL performance decreases when NNT uses patches of smaller or larger size than the ground truth, but it remains competitive. However, performance collapses if the patches are not quantized. 
    [Middle] Removing any rung of the ladder of improvements leads to a drop in performance.
    [Right] Warming up the world model before using it to train the policy on imaginary rollouts is required for good performance. BP denotes background planning. For each method, training in imagination
    starts at the color-coded vertical line, and leads to an initial drop in performance.}
    \label{fig:ablation}
\vspace{-.5em}
\end{figure*}

We conduct ablation studies to assess the importance of several components of our proposed MBRL agent.
Results are presented in Figure \ref{fig:ablation} and Table \ref{tab:ablation}. All the TWMs are trained for $N^{\text{iters}}_{\text{WM}}=500$ steps.

\textbf{Impact of patch size.} We investigate the sensitivity of our approach to the patch size used by NNT. While our best results are achieved when the tokenizer uses the oracle-provided ground truth patch size of $7\times 7$, Figure \ref{fig:ablation}[left] shows that performance remains competitive when using smaller ($5\times 5$) or larger ($9\times 9$) patches.

\paragraph{The necessity of quantizing.}
Figure \ref{fig:ablation}[left] shows that, when the $7\times7$ patches are not quantized, but instead the TWM is trained to reconstruct the continuous $7\times7$ patches, MBRL performance collapses. This is consistent with findings in DreamerV2 \citep{hafner2021benchmarking}, which highlight that quantization is critical for learning an effective world model.

\paragraph{Each rung matters.} To isolate the impact of each individual improvement, we remove each individual ``rung'' of our ladder from our best MBRL agent. As shown in Figure \ref{fig:ablation}[middle], each removal leads to a performance drop. This underscores the importance of combining all our proposed enhancements to achieve SOTA performance.

\paragraph{When to start training in imagination?} Training the policy on imaginary TWM rollouts requires a reasonably accurate world model. This is why background planning (Step 4 in Algorithm \ref{algo:MBRL})
only begins after $T_{\text{BP}}$ environment steps.
Figure \ref{fig:ablation}[right]
explores the effect of varying $T_{\text{BP}}$.
Initiating imagination training too early ($T_{\text{BP}}=0$) leads to performance collapse due to the inaccurate TWM dynamics.

\paragraph{MFRL ablation.} The final 3 rows in Table \ref{tab:ablation} show that either removing the RNN or using a smaller model as in \citet{moon2024discovering} leads to a drop in performance.


\paragraph{Annealing the number of policy updates.}  We linearly increase the number of policy updates on imaginary rollouts in Step 4 of Algorithm 1 from $N^{\text{iters}}_{\text{AC}}=0$ (when $T_{\text{total}}=0$) to $N^{\text{iters}}_{\text{AC}}=300$ (when $T_{\text{total}}=1$M). This annealing technique achieves a reward of $65.71\% (\pm 1.11)$, while removing the drop in performance observed when we start training in imagination. See Figure \ref{fig:annealing} Appendix \ref{sec:appendix_annealing}.

\subsection{Comparing TWM rollouts}
\label{sec:stationary}

In this section, we compare the TWM rollouts learned by three world models in our ladder, namely M1, M3 and our best model M5 (fast). 
To do so, we first create an evaluation dataset of $N_{\text{eval}}=160$ trajectories, each of length $T_{\text{eval}}=T_{\text{WM}}=20$, collected during the training of our best MFRL agent:
$\mathcal{D}_{\text{eval}} = \left\{O^{1:N_{\text{eval}}}_{1:T_{\text{eval}}+1}, a^{1:N_{\text{eval}}}_{1:T_{\text{eval}}}, r^{1:N_{\text{eval}}}_{1:T_{\text{eval}}} \right\}$.
We evaluate the quality of imagined trajectories generated by each TWM. 
Given a TWM checkpoint at 1M steps and the $n$th trajectory in $\mathcal{D}_{\text{eval}}$, we execute the sequence of actions $a^n_{1:T_{\text{eval}}}$, starting from $O^n_{1}$, to obtain a rollout trajectory $\hat{O}^{\text{TWM},~n}_{1:T_{\text{eval}} + 1}$. 


\begin{figure*}[t]
    \centering
    \begin{subfigure}[b]{0.3\textwidth}
        \centering
        \includegraphics[width=\linewidth]{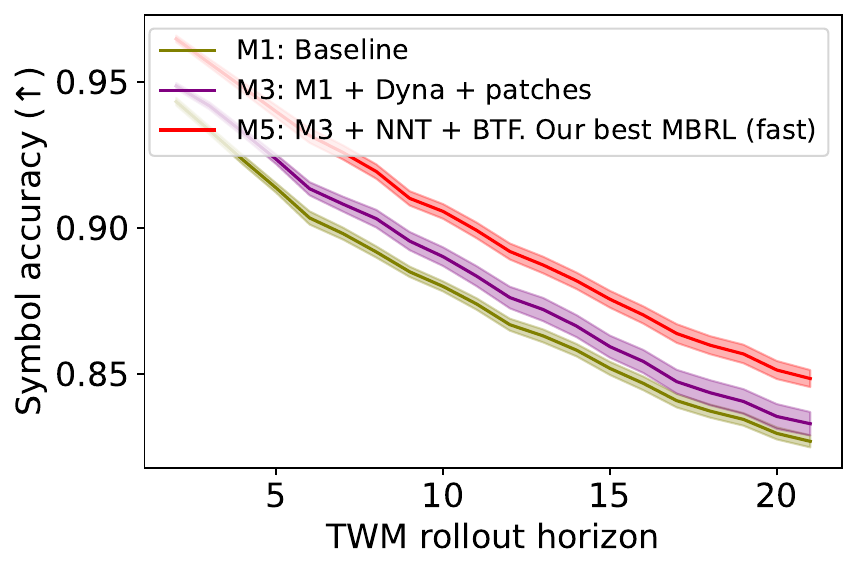}
        \hrule
        \vspace{0.3em}
        \includegraphics[width=0.8\linewidth]{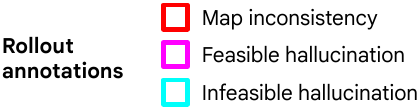}
    \end{subfigure}
    \hspace{0.5em}
    \begin{subfigure}[b]{0.65\textwidth}
        \includegraphics[width=\linewidth]{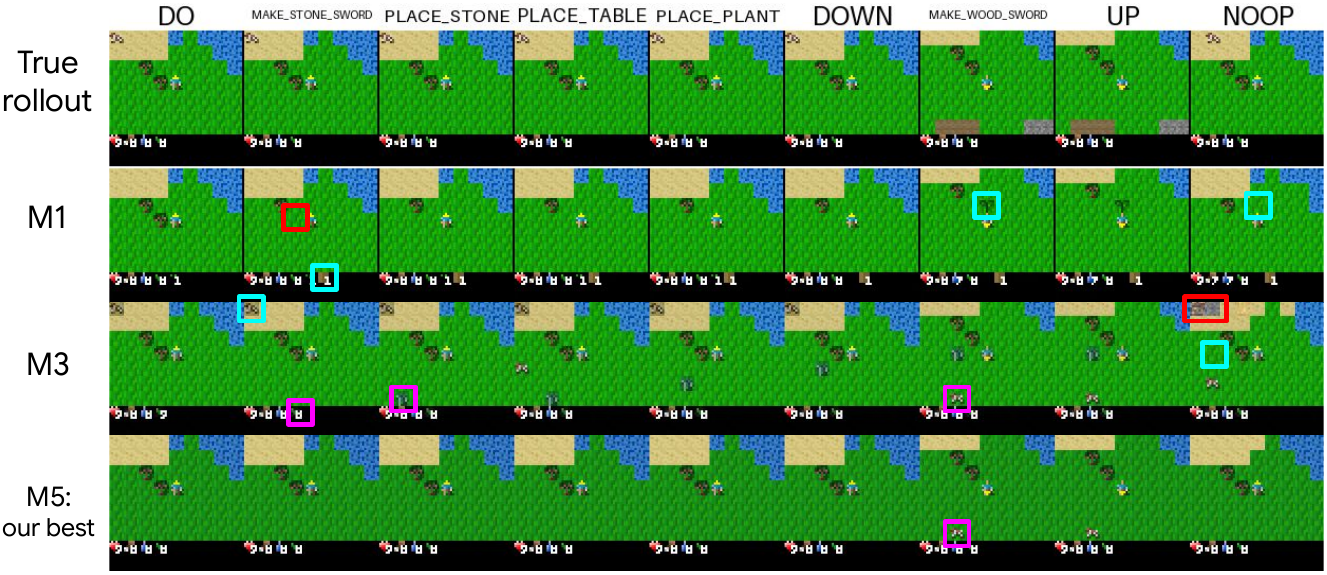}
    \end{subfigure}
    \vspace{-0.25em}
    \caption{
    Rollout comparison for world models M1, M3 and M5 (fast).
    [Left]
    Symbol accuracies decrease with the TWM rollout step. 
    The stationary NNT codebook used by M5 makes it easier to learn a reliable TWM.
    [Right]
    Best viewed zoomed in.
    \textbf{Map.}
    All three models accurately capture the agent's motion.
    All models can struggle to use the history to generate a consistent map when revisiting locations, however only M1 makes simple map errors in successive timesteps.
    \textbf{Feasible hallucinations.}
    M3 and M5 generate realistic hallucinations that respect the game dynamics, such as spawning mobs and losing health.
    \textbf{Infeasible hallucinations.}
    M1 often does not respect game dynamics; M1 incorrectly adds wood inventory, and incorrectly places a plant at the wrong timestep without the required sapling inventory.
    M3 exhibits some infeasible hallucinations in which the monster suddenly disappears or the spawned cow has an incorrect appearance.
    M5 rarely exhibits infeasible hallucinations.
    Figure \ref{fig:more_rollouts} in Appendix \ref{ap:rollout_comparison} shows more rollouts with similar behavior.
    }
    \label{fig:rollout_eval}
\vspace{-.25em}
\end{figure*}

\eat{
\textbf{Quantitative evaluations.}
To quantitatively evaluate the rollouts, in Figure \ref{fig:rollout_eval}[left] we plot the average L$2$ reconstruction error at each timestep $t$:
$
    \mathcal{E}_t = 
    \frac{1}{N_{\text{eval}}} \sum_{n=1}^{N_{\text{eval}}} \| \hat{O}^{\text{TWM},~n}_t - O^n_t \|_2^2,
    ~~~\forall t.
$
As expected, the error increases with $t$ as mistakes compound over the rollout.
Our best method, which uses NNT, achieves the lowest errors for all timesteps, highlighting that a stationary codebook makes learning simpler for the TWM, and leads to more accurate TWM rollouts.

We include two additional quantitative evaluations in Appendix \ref{ap:symbol_accuracy}. 
First, M5 achieves the lowest tokenizer reconstruction error in Figure \ref{fig:wm_quality_appendix}[left]. 
Second, we leverage the fact that each observation in Craftax-classic comes with a symbolic representation to compare symbol prediction accuracy from TWM rollouts in Figure \ref{fig:wm_quality_appendix}[right].
This metric further highlights that M5 best captures the game dynamics. 
}

\paragraph{Quantitative evaluations.}
For evaluation,
we leverage an appealing property of Craftax-classic: each observation $O_t$ comes with an array of ground truth symbols $S_t=(S_t^{1:R})$, with $R=145$. Given $100\text{k}$ pairs $(O_t, S_t)$, we train a CNN $f_{\mu}$, to predict the symbols from the observation; $f_{\mu}$ achieves a $99\%$ validation accuracy. Next, we use $f_{\mu}$ to predict the symbols from the generated rollouts. Figure \ref{fig:rollout_eval}[left] displays the average symbol accuracy at each timestep $t$:
\begin{equation*}
    \mathcal{A}_t = \frac{1}{N_{\text{eval}}R} \sum_{n=1}^{N_{\text{eval}}} \sum_{r=1}^{R}\mathbf{1}(f_{\mu}^r(\hat{O}^{\text{TWM},~n}_t), S^{r,n}_t), ~ \forall t,
\end{equation*}
where $\mathbf{1}(x,y) = 1 ~\text{iff.}~ x=y $ (and $0$ o.w.), $S^{r,n}_t$ denotes the ground truth $r$th symbol in the array $S^{n}_t$ associated with $O^n_t$, and $f_{\mu}^r(\hat{O}^{\text{TWM},~n}_t)$ its prediction for the rollout observation.
As expected, symbol accuracies decrease with $t$ as mistakes compound over the rollouts.
Our best method, which uses NNT, achieves the highest accuracies for all timesteps, as it best captures the game dynamics. This highlights that a stationary codebook makes TWM learning simpler.

We include two additional quantitative evaluations in Appendix \ref{ap:symbol_accuracy}, showing that M5 achieves the lowest tokenizer reconstruction errors and rollout reconstruction errors.

\paragraph{Qualitative evaluations.}
Due to environment stochasticity, TWM rollouts can differ from the environment rollout but still be useful for learning in imagination---as long as they respect the game dynamics.
Visual inspection of rollouts in Figure \ref{fig:rollout_eval}[right] reveals (a) map inconsistencies, (b) feasible hallucinations that respect the game dynamics and (c) infeasible hallucinations. 
M1 can make simple mistakes in both the map and the game dynamics. 
M3 and M5 both generate feasible hallucinations of mobs, however M3 more often hallucinates infeasible rollouts.

\eat{
\begin{figure}[t!]
    \centering
    \caption{MFRL ablations between ours and Moon.}
\end{figure}
}
\eat{
\begin{figure}[t!]
    \centering
    \caption{Sweep over $T_{WM}$ from 5-30. TWM sequence length is always 20. Link to rollout accuracy plot.}
\end{figure}
}

\subsection{Craftax Full}
\label{sec:craftax_full}

Table \ref{tab:craftax_full} compares the performance of various agents on
the full version of Craftax \citep{matthews2024craftax}, a significantly harder extension of Craftax-classic,
with more levels and achievements. While the previous SOTA agent reached $2.3\%$ reward (on symbolic inputs), our MFRL agent reaches
$4.63\%$ reward. Similarly, while the recent SOTA MBRL \citep{cohen2025text} reaches $6.59\%$ reward  our MBRL agent reaches a new SOTA reward of $7.20\%$.
See Appendix \ref{sec:classic_vs_full} for implementation details.

\begin{table}[h!]
\caption{
Results on Craftax after 1M environment interactions.
The previous SOTA scores are unknown.
}
\label{tab:craftax_full}
\vspace{-.5em}
\centering
\begin{tabular}{ccc}
\toprule
Method & Reward $(\%)$ & Score $(\%)$ \\
\midrule
Prev. SOTA MFRL  & $2.3$ (symbolic) & --- \\
Our best MFRL &  $4.63 \pm 0.20$ & $1.22 \pm 0.07$ \\
Prev. SOTA MBRL & $6.59$ & --- \\
Our best MBRL (slow)  & $\mathbf{7.20} \pm 0.09$ & $\mathbf{2.31} \pm 0.04$ \\
\bottomrule
\end{tabular}
\vspace{-.5em}
\end{table}

\subsection{Additional experiments on MinAtar}
\label{sec:minatar}
\begin{figure*}[b!]
\centering
\includegraphics[width=.95\linewidth]{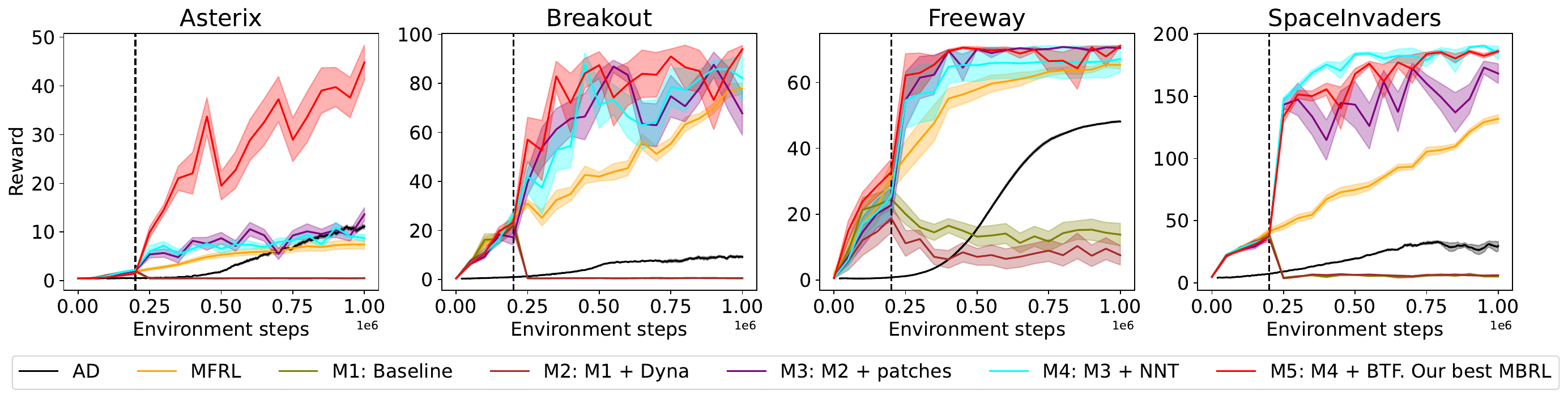}
\vspace{-.25em}
\caption{Our best MBRL agent outperforms our tuned MFRL agent on each MinAtar game.}
\label{fig:minatar_per_game}
\vspace{-.5em}
\end{figure*}

To further validate the robustness of our approach, we conduct additional experiments on MinAtar \citep{young19minatar}, another grid world environment. MinAtar implements four simplified Atari $2600$ games. Each game has symbolic binary observations of size $10\times10\times K$ ($K$ is the number of objects of the game) and binary rewards.

We first tune our model-free RL agent on the MinAtar games, keeping the same architecture as described in our paper, with minor adjustments to the PPO hyperparameters, detailed in Appendix \ref{sec:appendix_minatar}. Second, we develop our model-based RL agent as in Craftax-classic, by integrating our three proposed improvements. We retain the majority of the MBRL hyperparameters from Craftax-classic, with minor modifications, which we detail in Appendix \ref{sec:appendix_minatar}.

\begin{figure}[h!]
    \begin{minipage}{0.45\textwidth} 
    \centering
    \includegraphics[width=.95\linewidth]{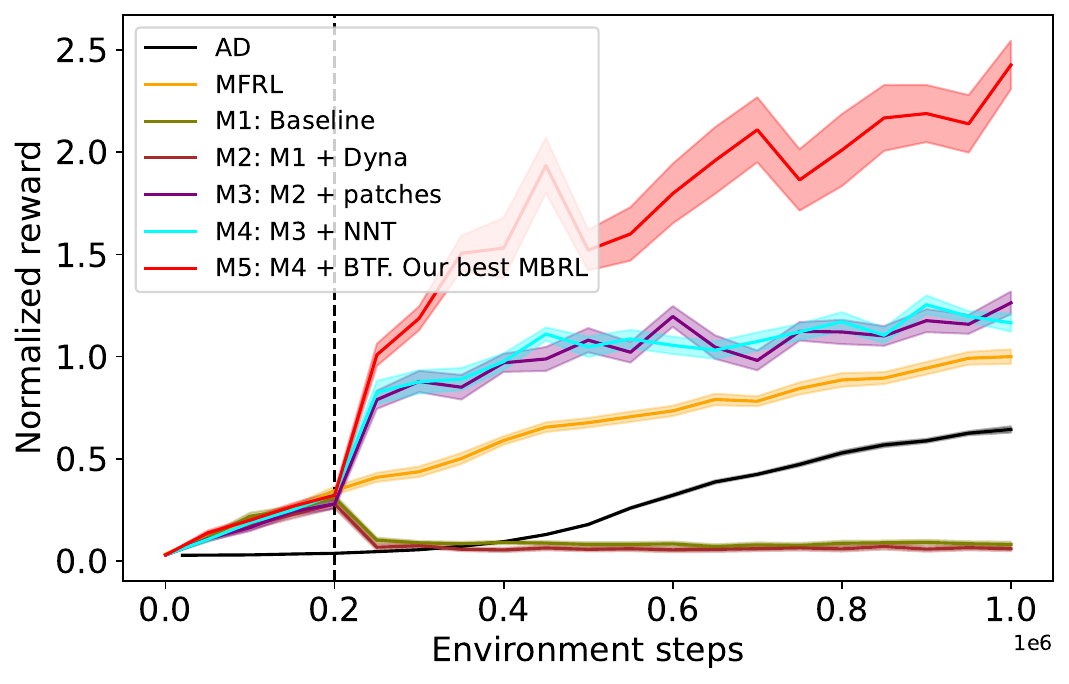}
    \vspace{-1em}
    \caption{Averaged normalized reward.}
    \label{fig:minatar_normalized_summary}
    \end{minipage}
    \hspace{0.4em}
    \begin{minipage}{0.45\textwidth} 
    \caption{
    Our MFRL and our best MBRL rewards after $1$M steps on each MinAtar game, compared with AD \citep{guan2024temporal} rewards after $5$M steps.
    }
    \label{tab:minatar}
    \centering
    \resizebox{\textwidth}{!}{
    \begin{tabular}{cccc}
    \toprule
    Game & MFRL $1$M & MBRL $1$M & AD $5$M \\
    \midrule
    Asterix & $7.47 \pm 1.02$ & $\mathbf{44.81} \pm 3.54$ & $21.05 \pm 0.65$\\
    Breakout &  $77.8 \pm 2.28$ & $\mathbf{93.92} \pm 1.44$  & $27.78 \pm 0.16$ \\
    Freeway & $65.3 \pm 1.16$ & $\mathbf{71.12} \pm 0.13$ & $57.68 \pm 0.07$  \\
    SpaceInvaders  & $131.9 \pm 3.32$ & $\mathbf{186.16} \pm 1.25$ & $140.36 \pm 1.70$ \\
    \bottomrule
    \end{tabular}
    }
    \end{minipage}
\end{figure}

Figure \ref{fig:minatar_per_game} displays the evaluation performance of our proposed methods M1-5 (defined as in Section \ref{sec:ladder}) on each game after $1$ million environment steps, averaged over $10$ seeds. Every $50$k training steps, we evaluate each agent on $32$ environments and $2$k steps per environments.
We compare our methods to the recent Artificial Dopamine agent of \citep{guan2024temporal}---referred to as AD---using the results shared by the authors.
Figure \ref{fig:minatar_normalized_summary} summarizes these results by first (a) normalizing each game such that the MFRL agent achieves a reward of $1.0$, before (b) averaging the performance of all agents across the games.
Notably, our MBRL agents' performance increase as we climb the ladder on MinAtar, highlighting the generality of our three proposed improvements. Furthermore, our best MBRL agent significantly outperforms our best MFRL agent, achieving an average normalized reward of $2.43$ across the four MinAtar games. In contrast, the AD agent reaches an average normalized reward of $0.64$, highlighting the performance of our tuned MFRL agent. 

Finally, Table \ref{tab:minatar} compares the performance of our best MBRL and MFRL agents at $1$M steps, with the AD agent at $5$M steps, further emphasizing the significant performance improvements achieved by our proposed MBRL agent.

\subsection{Extensions to multiplayer games}
\label{sec:openspiel}

Finally, we extend our framework to encompass three two-player zero-sum board games from the OpenSpiel suite \citep{LanctotEtAl2019OpenSpiel}: Bargaining \citep{lewis2017deal}, Leduc Poker \citep{southey2012bayes}, and Tic-Tac-Toe. Tic-Tac-Toe is fully observed (all information is available to both players) and has deterministic dynamics. In contrast, Bargaining and Leduc Poker are partially observed (players have incomplete information about their opponent's observations) and have stochastic dynamics. Our goal is to train a single agent (either Player 1 or Player 2) to maximize its reward when competing against an opponent that uniformly picks a legal action. Extending MBRL to these multiplayer games is particularly challenging, as the TWM must accurately simulate the game dynamics, accounting for actions from both players and any chance events.

Observations in OpenSpiel are represented as a sequence of symbols, so we do not need NNT (nearest-neighbor tokenizer). We make two additional modeling choices. First, to make TWM training easier, we convert stochastic dynamics into deterministic dynamics. To do so, we introduce a ``chance player'' which takes discrete ``chance actions'' (e.g. rolling a dice), which are distinct from players' actions. Second, we assume that, unlike the policy, the world model is trained on fully visible observations. Specifically, when collecting data from the environment (Step 1 of Appendix \ref{algo:MBRL}), we gather (a) the current player ID ($0$ for the chance player, $1$ for Player, $2$ for Player), (b) both players' observations, (c) the list of legal actions for the current player (including their probabilities for the chance player), (d) the action taken by the current player.

Based on the history of observations and actions, the TWM is trained to predict the reward, termination, and the next observations for both players. Additionally, it predicts (a) the next player ID, (b) the next set of legal actions for both players, and (c) the probabilities of the next chance actions. 
This specific training enables the TWM to produce imaginary rollouts that respect the game's rules and accurately simulate the interplay of players and chance actions.
During policy training in imagination (Step 4 of Appendix \ref{algo:MBRL}), we extract only our agent's predicted observation sequence, and discard the opponent's predicted observations.

Our MFRL agent uses the same hyperparameters as Craftax-classic. For MBRL, we reuse the hyperparameters from Minatar. See Appendix \ref{sec:appendix_twm_multiplayer} for details. Due to the simplicity of the OpenSpiel games, we use $T_{\text{total}}=100\text{k}$ actions in total for both players, and start training in imagination at $T_{\text{BP}}=0$ steps. 

\begin{figure*}[h!]
\centering
\includegraphics[width=.95\linewidth]{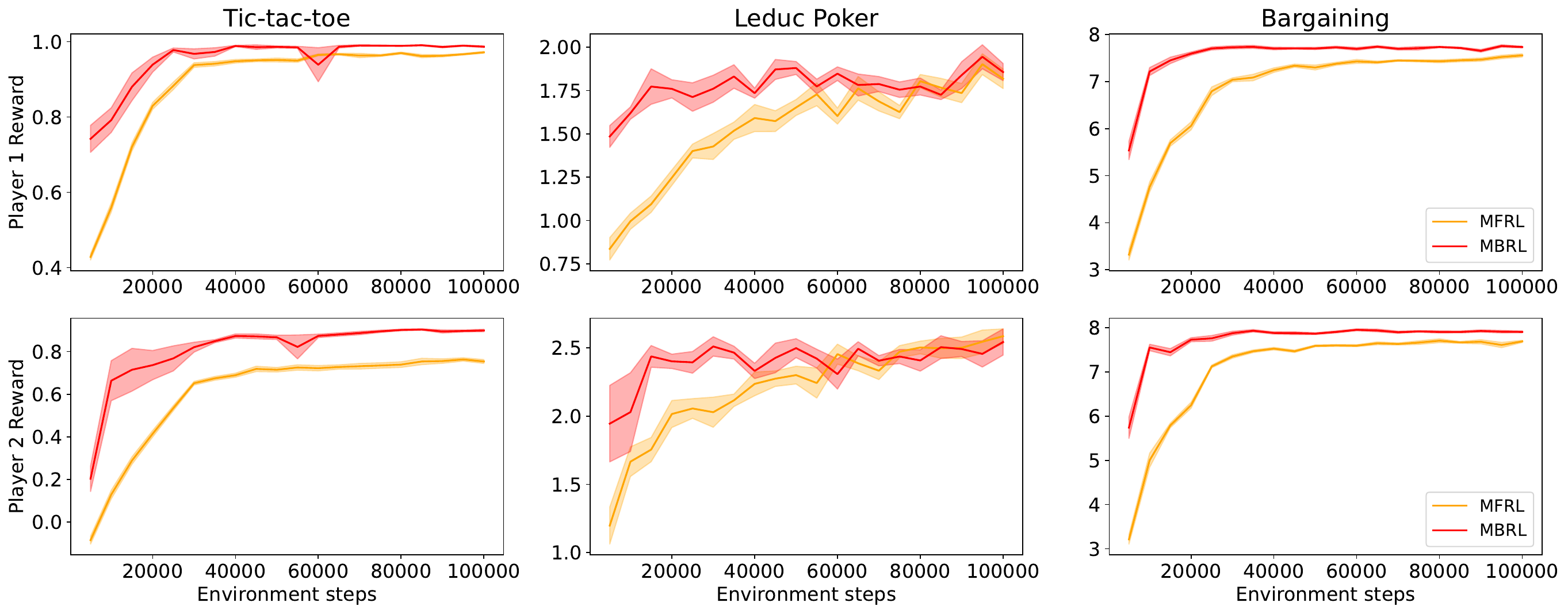}
\vspace{-.25em}
\caption{By leveraging a multiplayer TWM, our MBRL agents achieve higher rewards than our MFRL agents across three two-player games, illustrating the generality of our approach.}
\label{fig:openspiel_per_game}
\vspace{-.5em}
\end{figure*}

Figure \ref{fig:openspiel_per_game} compares our MFRL and MBRL agents on each game, averaged over $10$ seeds, when playing against an opponent that uniformly picks a legal action. Notably, our MBRL agents consistently achieve higher rewards than our MFRL agents, with this performance gap being particularly pronounced in the early stages of training. These results further highlight the broad applicability of our framework and demonstrate that, for simple symbolic games, a TWM can be effectively learned from as little as tens of thousands of environment interactions and used to train a MBRL agent in imaginary rollouts.

%% file: arxiv/concl.tex
\section{Conclusion and future work}
In this paper, we present three
improvements to vision-based MBRL agents which use transformer
world models for background planning: Dyna with warmup,
patch nearest-neighbor tokenization and \btf. 
We also present improvements to the MFRL baseline, which may be of independent interest.
Collectively, these improvements result in a MBRL agent that achieves a significantly higher reward and score than previous SOTA agents on the challenging Craftax-classic benchmark, and surpasses expert human reward for the first time. Our improvements also transfer to Craftax-full, MinAtar environments,
and three different two-player games.
In the future, we plan to examine how well our techniques generalize
beyond grid-world environments.
However, we believe our current results will already be of interest to the community.

We see several paths to build upon our method.
Prioritized experience replay is a promising approach to accelerate TWM training, and an off-policy RL algorithm could improve policy updates by mixing imagined and real data.
In the longer term, we would like to generalize our tokenizer to extract
patches and tokens from large pre-trained models,
such as SAM \citep{ravi2024sam} and Dino-V2 \citep{Oquab2024}.
This inherits the stable codebook of our approach, but reduces sensitivity to patch size and ``superficial" appearance variations.
To explore this direction, and other non-reconstructive world models which cannot generate future pixels,  we plan to modify the policy to directly accept latent tokens generated by the TWM.

\bigskip

\section*{Acknowledgments}
We thank Pablo Samuel Castro for useful discussions during the preparation of this manuscript.

%% file: arxiv/appendix.tex
\newpage
\appendix
\onecolumn

\section{Algorithmic details}
\label{sec:hyperparameters}

\subsection{Our Model-free RL agent}
\label{ap:mfrl_agent}

We first detail our new state-of-the-art MFRL agent. As mentioned in the main text, it relies on an actor-critic policy network trained with PPO.

\subsubsection{MFRL architecture}

We summarize our MFRL agent in Algorithm \ref{algo:ac_network} and further detail it below.
\begin{algorithm}[h!]
\caption{MFRL agent}
\label{algo:ac_network}
\begin{algorithmic}
\STATE {\bfseries Input:} Image $O_t$, last hidden state $h_{t-1}$, parameters $\Phi$. 
\STATE {\bfseries Output:} action $a_t$, value $v_t$, new hidden state $h_t$.
\STATE $z_t = \text{ImpalaCNN}_{\Phi}(O_t)$
\STATE $h_t, y_t = \text{RNN}_{\Phi}( [h_{t-1}, z_t])$
\STATE $a_t \sim \pi_{\Phi}([y_t, z_t])$
\STATE $v_t = V_{\Phi}([y_t, z_t])$
\end{algorithmic}
\end{algorithm}

\textbf{Imapala CNN architecture: }
Each Craftax-classic image $O_t$ of size $63\times63\times3$ goes through an Impala CNN \citep{espeholt2018impala}.
The CNN consists of three stacks with channel sizes of ($64, 64, 128$).  Each stack is composed of (a) a batch normalization \citep{ioffe2015batch}, (b) a convolutional layer with kernel size $3\times3$ and stride of $1$, (c) a max pooling layer with kernel size $3\times3$ and stride of $2$, and (d) two ResNet blocks \citep{he2016deep}. Each ResNet block is composed of (a) a ReLU activation followed by a batch normalization, (b) a convolutional layer with kernel size $3\times3$ and stride of $1$. The CNN last layer output, of size $8 \times 8 \times 128$ passes through a ReLU activation, then gets flattened into an embedding vector of size $8192$, which we call $z_t$.

\paragraph{RNN architecture:} The CNN output $z_t$ (a) goes through a layer norm operator, (b) then gets linearly mapped to a $256$-dimensional vector, (c) then passes through a ReLU activation, resulting in the new input for the RNN. The RNN then updates its hidden state, and outputs a $256$-dimensional vector $y_t$, which goes through another ReLU activation.

\paragraph{Actor and critic architecture:} Finally, the CNN output $z_t$ and the RNN output $y_t$ are concatenated, resulting in the $8448$-dimensional embedding input shared by the actor and the critic networks. For the actor network, this shared input goes through (a) a layer normalization \citep{lei2016layer}, (b) a fully-connected network whose $2048$-dimensional output goes through a ReLU, (c) two dense residual blocks whose $2048$-dimensional output goes through a ReLU, (d) a last layer normalization and (e) a final fully-connected network which predicts the action logits.

Similarly, for the critic network, the shared input goes through (a) a layer normalization, (b) a fully-connected network whose $2048$-dimensional output goes through a ReLU, (c) two dense residual blocks whose $2048$-dimensional output goes through a ReLU, (d) a last layer normalization and (e) a final layer which predicts the value (which is a float).

\subsubsection{PPO training}
We train our MFRL agent with the PPO algorithm \citep{schulman2017proximal}. PPO is a policy gradient algorithm, which we briefly summarize below.

\paragraph{Training objective:}
We assume we are given a trajectory, $\tau=(O_{1:T+1}, a_{1:T}, r_{1:T}, \text{done}_{1:T}, h_{0:T})$ collected in the environment, where $\text{done}_t$ is a binary variable indicating whether the current state is a terminal state, and $h_t$ is the RNN hidden state collected while executing the policy. Algorithm \ref{algo:rollout} discusses how we collect such a trajectory.

Given the fixed current actor-critic parameters $\Phi_{\text{old}}$,
PPO first runs the actor-critic network on $\tau$, starting from the hidden state $h_0$ and returns two sequences of values $v_{1:T+1}=V_{\Phi_{\text{old}}}(O_{1:T+1})$ and probabilities $\pi_{\Phi_{\text{old}}}(a_t | O_t)$\footnote{We drop the ImpalaCNN and the RNN for simplicity.}. It then defines the generalized advantage estimation (GAE) as in \citet{schulman2015high}:
\begin{equation*}
A_t = \delta_t + (1 - \text{done}_t)  \gamma \lambda A_{t+1} = \delta_t + (1 - \text{done}_t)  \left(\gamma \lambda \delta_{t+1} + \ldots + (\gamma \lambda)^{T-t} \delta_{T}\right). ~~~~ \forall t \le T \\
\end{equation*}
where
$$
\delta_t = r_t + (1 - \text{done}_t)  \gamma v_{t+1} - v_t.
$$
PPO also defines the TD targets $q_t = A_t + v_t$.

PPO optimizes the parameters $\Phi$, to minimize the objective value:
\begin{equation}\label{ppo}
\mathcal{L}_{\text{PPO}}(\Phi)
=
\frac{1}{T} \sum_{t=1}^T \left\{ -\min \left(r_t(\Phi) A_t, \text{clip}(r_t(\Phi)) A_t) \right)
+ \lambda_{\text{TD}} (V_{\Phi}(O_t) - q_t)^2
- \lambda_{\text{ent}} \mathcal{H}(\pi_{\Phi}(. | O_t)) \right\},
\end{equation}
where 
$\text{clip}(u)$ ensures that $u$ lies in the interval $[1-\epsilon, 1+\epsilon]$,
$r_t(\Phi)$ is the probability ratio $r_t(\Phi) = \frac{\pi_{\Phi}(a_t | O_t)}{ \pi_{\Phi{\text{old}}}(a_t | O_t) }$ and $\mathcal{H}$ is the entropy operator. 


\paragraph{Algorithm:}
Algorithm \ref{algo:ppo_update} details the PPO-update-policy, which is called in Steps 1 and 4 in our main Algorithm \ref{algo:MBRL} to update the PPO parameters on a batch of trajectories. PPO allows multiple epochs of minibatch updates on the same batch and introduces two hyperparameters: a number of minibatches $N^{\text{mb}}$ (which divides the number of environments $N_{\text{env}}$), and a number of epochs $N^{\text{epoch}}$.

\medskip
\begin{algorithm}[h!]
\small
\caption{PPO-update-policy}
\label{algo:ppo_update}
\begin{algorithmic}
\STATE {\bfseries Input:} Actor-critic model $(\pi, V)$ and parameters $\Phi$
\newline
Trajectories $\tau^{1:N_{\text{env}}}=(O^{1:N_{\text{env}}}_{1:T+1}, a^{1:N_{\text{env}}}_{1:T}, r^{1:N_{\text{env}}}_{1:T}, \text{done}^{1:N_{\text{env}}}_{1:T}, h^{1:N_{\text{env}}}_{0:T})$
\newline
Number of epochs $N^{\text{epoch}}$ and of minibatches $N^{\text{mb}}$
\newline
PPO objective value parameters $\gamma, \lambda, \epsilon$
\newline
Learning rate lr and max-gradient-norm
\newline
Moving average mean $\mu_{\text{target}}$, standard deviation $\sigma_{\text{target}}$ and discount factor $\alpha$
\medskip
\STATE {\bfseries Output:} Updated actor-critic parameters $\Phi$
\medskip
\STATE {\bfseries Initialize:} Define $\Phi_{\text{old}} = \Phi$ 
\STATE Compute the values $v^{1:N_{\text{env}}}_{1:T+1} = V_{\Phi_{\text{old}}}(O^{1:N_{\text{env}}}_{1:T+1})$
\STATE Compute PPO GAEs and targets $A^{1:N_{\text{env}}}_{1:T}, q^{1:N_{\text{env}}}_{1:T} = \text{GAE}(r^{1:N_{\text{env}}}_{1:T}, v^{1:N_{\text{env}}}_{1:T+1}, \gamma, \lambda)$
\STATE Standardize PPO GAEs $A^{1:N_{\text{env}}}_{1:T} = \frac{A^{1:N_{\text{env}}}_{1:T} - \text{mean}(A^{1:N_{\text{env}}}_{1:T})}{\text{std}(A^{1:N_{\text{env}}}_{1:T})}$
\medskip
\FOR{$\text{ep}=1$ {\bfseries to} $N^{\text{epoch}}$}
\FOR{$k=1$ {\bfseries to} $N^{\text{mb}}$}
\STATE $N^{\text{start}} = (k-1) \left(N_{\text{env}} / N^{\text{mb}}\right) +1, ~~ N^{\text{end}} = k\left(N_{\text{env}} / N^{\text{mb}}\right) +1$
\STATE \textit{// Standardize PPO target}
\STATE Update $\mu_{\text{target}} = \alpha \mu_{\text{target}} + (1-\alpha) \text{mean}(q^{N^{\text{start}}:N^{\text{end}}}_{1:T})$
\STATE Update $\sigma_{\text{target}} = \alpha \sigma_{\text{target}} + (1-\alpha) \text{std}(q^{N^{\text{start}}:N^{\text{end}}}_{1:T})$
\STATE Standardize $q^{N^{\text{start}}:N^{\text{end}}}_{1:T} = (q^{N^{\text{start}}:N^{\text{end}}}_{1:T} - \mu_{\text{target}}) / \sigma_{\text{target}}$

\medskip
\STATE \textit{// Run the actor-critic network}
\STATE Define $\tilde{h}^{N^{\text{start}}:N^{\text{end}}}_0= h^{N^{\text{start}}:N^{\text{end}}}_0$
\FOR{$t=1$ {\bfseries to} $T+1$}
\STATE $z^n_t = \text{ImpalaCNN}_{\Phi}(O^n_{t}) ~~~; ~~~\tilde{h}^n_t = \text{RNN}_{\Phi}( [\tilde{h}^n_{t-1}, z^n_t])$ 
~~~~~~~~~~~~~~~~~~~~~ for $n=N^{\text{start}}:N^{\text{end}}$
\STATE Compute $V^n_{\Phi}([y^n_t, z^n_t])$ and $\pi^n_{\Phi}([y^n_t, z^n_t])$ 
~~~~~~~~~~~~~~~~~~~~~~~~~~~~~~~~~~~~~~~~~~~~ for $n=N^{\text{start}}:N^{\text{end}}$
\ENDFOR

\medskip
\STATE \textit{// Take a gradient step}
\STATE Compute $\mathcal{L}^n_{\text{PPO}}(\Phi)$ using Equation \eqref{ppo}
~~~~~~~~~~~~~~~~~~~~~~~~~~~~~~~~~~~~~~~~~~~~~~~~~~~~~ for $n=N^{\text{start}}:N^{\text{end}}$
\STATE Define the minibatch loss $\mathcal{L}_{\text{PPO}}(\Phi) = \frac{1}{N^{\text{mb}}} \sum_{n=N^{\text{start}}}^{N^{\text{end}}} \mathcal{L}^n_{\text{PPO}}(\Phi) $
\STATE Update $\Phi = \text{Adam}
\left(\Phi,
~~ \text{clip-gradient}(\nabla_{\Phi} \mathcal{L}_{\text{PPO}}(\Phi), \text{max-norm)}, ~~\text{lr} \right)$
\ENDFOR
\ENDFOR
\end{algorithmic}
\end{algorithm}

We make a few comments below:

$\bullet$ We use gradient clipping on each minibatch to control the maximum gradient norm, and update the actor-critic parameters using Adam \citep{kingma2014adam} with learning rate of $0.00045$. 

$\bullet$ During each epoch and minibatch update, we initialize the hidden state $\tilde{h}_0$ from its value $h_0$ stored while collecting the trajectory $\tau$.

$\bullet$ In Algorithm \ref{algo:ppo_update}, we introduce two changes to the standard PPO objective, described in Equation \eqref{ppo}.
First, we standardize the GAEs
(ensure they are zero mean and unit variance)
across the batches. Second, similar to \citet{moon2024discovering}, we maintain a moving average with discount factor $\alpha$ for the mean
and standard deviation of the target $q_t$ and we update
the value network to predict the standardized targets.

\paragraph{Implementation: }
Note that for implementing PPO, we start from the code available in the \texttt{purejaxrl} library \citep{lu2022discovered} at \url{https://github.com/luchris429/purejaxrl/blob/main/purejaxrl/ppo.py}.

\subsubsection{Hyperparameters}
Table \ref{table:mfrl_hyperparameters} displays the PPO hyperparameters used for training our SOTA MFRL agent.

\begin{table*}[h!]
\small
\centering
\caption{MFRL hyperpameters.}
\label{table:mfrl_hyperparameters}
\begin{tabular}{p{0.2\textwidth} p{0.35\textwidth}p{0.3\textwidth} }
    \toprule
    Module & Hyperparameter & Value\\
    \toprule
    \toprule
    Environment & Number of environments $N_{\text{env}}$ & $48$ \\
    & Rollout horizon in environment $T_{\text{env}}$ & $96$ \\
    \midrule
    Sizes & Image size & $63\times63\times3$ \\
    & CNN output size & $8\times8\times128$ \\
    & RNN hidden layer size & $256$ \\
    & AC input size & $8448$ \\
    & AC layer size & $2048$ \\
    \midrule
    PPO & $\gamma$ & $0.925$ \\
    & $\lambda$ & $0.625$ \\
    & $\epsilon$ clipping & $0.2$ \\
    & TD-loss coefficient $\lambda_{\text{TD}}$ & $1.0$ \\
    & Entropy loss coefficient $\lambda_{\text{ent}}$ & $0.01$ \\
    & PPO target discount factor $\alpha$ & $0.95$ \\
    \midrule
    Learning 
    & Optimizer & Adam \citep{kingma2014adam}\\
    & Learning rate & $0.00045$ \\
    & Max. gradient norm & $0.5$ \\
    & Learning rate annealing (MFRL) & True (linear schedule) \\
    & Number of minibatches (MFRL) & $8$ \\
    & Number of epochs (MFRL) & $4$ \\
\bottomrule
\end{tabular}
\end{table*}

\textbf{MBRL experiments.}
We make two additional changes to PPO in the MBRL setting, and keep all the other hyperparameters fixed.
First,
we do not use learning rate annealing for MBRL,
 while MFRL uses learning rate annealing (with a linear schedule).
 Second, as we discuss in Section \ref{sec:minibatch_epoch}, the differences between the PPO updates on real and imaginary trajectories lead to varying the number of minibatches and epochs.

\paragraph{Craftax experiments.} We also keep all but two of our PPO hyperparameters fixed for Craftax (full), which we discuss in Appendix \ref{sec:classic_vs_full}.

\newpage
\subsection{Model-based modules}
\label{ap:mbrl_modules}

In this section, we detail the two key modules for model-based RL: the tokenizer and the transformer world model.

\subsubsection{Tokenizer}\label{sec:appendix_ae}

\paragraph{Training objective: }
Given a Craftax-classic image $O_t$ and a codebook $\mathcal{C} = \{e_1, \ldots, e_K\}$, an encoder $\mathcal{E}$ returns a feature map $Z_t = (Z_t^1, \ldots, Z^t_L)$. Each feature $Z_t^{\ell}$ gets quantized, resulting into $L$ tokens $Q_t = (q_t^1,\ldots, q_t^L)$---which serves as input to the TWM---then projected back to $\hat{Z}_t = (e_{q_t^1}, \ldots e_{q_t^L})$. Finally, a decoder $\mathcal{D}$ decodes $\hat{Z}_t$ back to the image space: $\hat{O}_t = \mathcal{D}(\hat{Z}_t)$. Following \citet{van2017neural,micheli2022transformers}, we define the VQ-VAE loss as:
\begin{equation}\label{vqvae}
\mathcal{L}_{\text{VQ-VAE}}(\mathcal{E}, \mathcal{D}, \mathcal{C}) =
\lambda_1 \|O_t - \hat{O}_t \|_1 
+ \lambda_2 \|O_t - \hat{O}_t \|_2^2 
+ \lambda_3 \| \text{sg}(Z_t) - \hat{Z}_t \|_2^2
+ \lambda_4 \| Z_t - \text{sg}(\hat{Z}_t) \|_2^2,
~.
\end{equation}
where $\text{sg}$ is the stop-gradient operator. 
The first two terms are the reconstruction loss, the third term is the codebook loss and the last term is a commitment loss.

\medskip
We now discuss the different VQ and VQ-VAE architectures used by the models M1-5 in the ladder described in Section \ref{sec:ladder}.

\paragraph{Default VQ-VAE:} Our baseline model M1, and our next model M2 build on IRIS VQ-VAE \citep{micheli2022transformers} and follow the authors' code: \url{https://github.com/eloialonso/iris/blob/main/src/models/tokenizer/nets.py}. The encoder uses a convolutional layer (with kernel size $3\times3$ and stride $1$), then five residual blocks with two convolutional layers each (with kernel size $3\times3$, stride $1$ and ReLU activation). The channel sizes of the residual blocks are $(64, 64, 128, 128, 256)$. A downsampling is applied on the first, third and fourth blocks. Finally, a last convolutional layer with $128$ channels returns an output of size $8 \times 8 \times 128$.
The decoder follows the reverse architecture. Each of the $L=64$ latent embeddings gets quantized individually, using a codebook of size $K=512$, to minimize Equation \eqref{vqvae}. 
We use codebook normalization, meaning that each code in the codebook $\mathcal{C}$ has unit L$2$ norm. Similarly, each latent embedding $Z_t^{\ell}l$ gets normalized before being quantized. As in IRIS, we use $\lambda_1=1, \lambda_2=0, \lambda_3=1, \lambda_4=0.25$. 
We train with Adam and a learning rate of $0.001$.

\paragraph{VQ-VAE(patches):} For the next model M3, the encoder is a two-layers MLP that maps each flattened $7\times 7 \times 3$ patch to a $128$-dimensional vector, using a ReLU activation. Similarly, the decoder learns a linear mapping from the embedding vector back to the flattened patches. Each embedding gets quantized individually, using a codebook of size $K=512$, and codebook normalization, to minimize Equation \eqref{vqvae}. Following \citet{micheli2024efficient}, we use $\lambda_1=0.1, \lambda_2=1, \lambda_3=1, \lambda_4=0.02$.

\paragraph{Nearest neighbor tokenizer:} NNT does not use Equation \eqref{vqvae} and directly adds image patches to a codebook of size $K=4096$, using a Euclidean threshold $\tau=0.75$.

\subsubsection{Transformer world model}\label{sec:appendix_twm}

\paragraph{Training objective:}
We train the TWM on real trajectories (from the environment) of $T_{\text{WM}}=20$ timesteps sampled from the replay buffer (see Algorithm \ref{algo:MBRL}). 
We set $T_{\text{WM}}=20$ as it is the largest value that will fit into memory on 8 H100 GPUs.

Given a trajectory $\tau=(O_{1:T+1}, a_{1:T}, r_{1:T}, \text{done}_{1:T})$, the input to the transformer is the sequence of tokens:
$$
(q_1^1,\ldots,q_1^L,a_1, \ldots q_T^1,\ldots,q_T^L, a_T),
$$
where $\text{enc}(O_t)=(q_t^1,\ldots,q_t^L)$ and $q_t^i \in \{1,\ldots,K\}$ where $K$ is the codebook size. 
These tokens are then embedded using an observation embedding table and an action embedding table. 
After several self-attention layers (using the standard causal mask or the block causal mask introduced in Section \ref{sec:btf}), the TWM returns a sequence of output embeddings:
$$
(E(q_1^1),\ldots,E(q_1^L),E(a_1), \ldots E(q_T^T),\ldots,E(q_1^T), E(a_T)).
$$
The TWM then output embeddings are then used to decode the following predictions:

\smallskip
$\textbf{(1)}$ Following \citep{micheli2022transformers}, $E(a_t)$ passes through a reward head and predicts the logits of the reward $r_t$.

\smallskip
$\textbf{(2)}$ $E(a_t)$ also passes through a termination head and predicts the logits of the termination state $\text{done}_t$. 

\smallskip
$\textbf{(3)}$ Without block teacher forcing, $E(q_t^i)$ passes through an observation head and predicts the logits of the next codebook index at the same timestep $E(q_t^{i+1})$, when $t \le L-1$. Similarly $E(a_t)$ passes through an observation head and predicts the logits of the first codebook index at the next timestep $E(q_{t+1}^{1})$.

\smallskip
$\textbf{(3')}$ With block teacher forcing, $E(q_t^i)$ passes through an observation head and predicts the logits of the same codebook index at the next timestep $E(q_{t+1}^i)$.

\medskip

TWM is then trained with three losses:

\smallskip
$\textbf{(1)}$
The first loss is the cross-entropy for the reward prediction. Note that Craftax-classic provides a (sparse) reward of 1 for the first time each achievement is``unlocked'' in each episode.
In addition, it gives a smaller (in magnitude) but denser reward, penalizing the agent by $0.1$ for every
point of damage taken, and rewarding it by $0.1$ for every point recovered. 
However, we found that
we got better results by ignoring the points damaged and recovered,
and using a binary reward target.
This is similar to the recommendations in 
\citet{Farebrother2024}, where the authors show that value-based RL methods work better when replacing MSE loss for value functions with cross-entropy on a quantized version of the return.

\smallskip
$\textbf{(2)}$
The second loss is the cross-entropy for the termination predictions. 

\smallskip
$\textbf{(3}$ 
The third loss is the cross-entropy for the codebook predictions, where the predicted codes vary between $1$ and the codebook size $K$.

\bigskip

\paragraph{Architecture:} We use the standard GPT2 architecture \citep{radford2019language}. We use a sequence length $T_{\text{WM}}=20$ due to memory constraints.
We implement key-value caching to generate rollouts fast. Table \ref{table:twm_hyperparameters} details the different hyperparameters.
\begin{table*}[h!]
\small
\centering
\caption{Hyperparameters for the transformer world model.}
\label{table:twm_hyperparameters}
\begin{tabular}{p{0.2\textwidth}p{0.35\textwidth}p{0.2\textwidth} }
\toprule
Module & Hyperparameter & Value\\
\toprule
\toprule
Environment & Sequence length $T_{\text{WM}}$ & $20$ \\
\midrule
Architecture  & Embedding dimension & $128$ \\
& Number of layers & $3$ \\
& Number of heads & $8$ \\
& Mask & Causal or Block causal\\
& Inference with key-value caching & True \\
& Positional embedding & RoPE \citep{su2024roformer} \\
\midrule
Learning & Embedding dropout & $0.1$ \\
& Attention dropout & $0.1$ \\
&  Residual dropout & $0.1$ \\
& Optimizer & Adam \citep{kingma2014adam} \\
& Learning rate & $0.001$ \\
& Max. gradient norm & $0.5$ \\
\bottomrule
\end{tabular}
\end{table*}


\newpage
\subsection{Our Model-based RL agent}
\label{ap:mbrl_agent}

In this section, we detail how we combine the different modules above to build our SOTA MBRL agent, which is described in Algorithm \ref{algo:MBRL} in the main text.

\subsubsection{Collecting environment rollout or TWM rollout}
Algorithm \ref{algo:rollout} presents the rollout method, which we call in Steps 1 and 4 of Algorithm \ref{algo:MBRL}. It requires a transition function which can either be the environment or the TWM.

\begin{algorithm}[h!]
\small
\caption{~Environment rollout or TWM rollout}
\label{algo:rollout}
\begin{algorithmic}
\STATE {\bfseries Input:} Initial observation $O_1$,
\newline
Previous $M$ observations $O_{\text{past}} = (O_{-M+1}, \ldots, O_0)$ if available else $O_{\text{past}} = \emptyset$,
\newline
AC model $\pi$ and parameters $\Phi$,
\newline
Rollout horizon $T$,
\newline
An environment transition $\Mtrue$ or a TWM $\mathcal{M}$ with parameters $\Theta$.
\medskip
\STATE {\bfseries Output:} A trajectory $\tau=(O_{1:T+1}, a_{1:T}, r_{1:T}, \text{done}_{1:T}, h_{0:T})$

\medskip
\STATE {\bfseries Initialize:} hidden state $h_0 =0$ if $O_{\text{past}} = \emptyset$ else set $h_{-M} =0$
\IF{$O_{\text{past}} \ne \emptyset$}
\STATE \textit{// Burn-in the hidden state}
\FOR{$m=1$ {\bfseries to} $M$}
\STATE $z_{-M + m} = \text{ImpalaCNN}_{\Phi}(O_{-M + m})$
\STATE $h_{-M + m} = \text{RNN}_{\Phi}( [h_{-M -1 + m}, z_{-M + m}])$
\ENDFOR
\ENDIF

\medskip
\STATE {\bfseries Initialize:} $\tau = (h_0)$

\medskip
\FOR{$t=1$ {\bfseries to} $T$}
\STATE \textit{// Run the actor network}
\STATE $z_t = \text{ImpalaCNN}_{\Phi}(O_t)$
\STATE $h_t = \text{RNN}_{\Phi}( [h_{t-1}, z_t])$
\STATE $a_t \sim \pi_{\Phi}([h_t, z_t])$
\medskip
\STATE \textit{// Collect reward and next observation}
\IF{~environment rollout~}
\STATE $O_{t+1}, r_t, \text{done}_t \sim \Mtrue(O_t, a_t)$
\ELSIF{~TWM rollout~}
\STATE $Q_t = (q_t^1, \ldots, q_t^L) = \text{enc}(O_t)$
\STATE $Q_{t+1} \sim p_{\Theta}(Q_{t+1} | Q_{1:t}, a_{1:t})$
\STATE $O_{t+1} = \text{dec}(Q_{t+1})$
\STATE $r_t \sim p_{\Theta}(r_t | Q_{1:t}, a_{1:t})$
\STATE $\text{done}_t \sim p_{\Theta}(\text{done}_t | Q_{1:t}, a_{1:t})$
\ENDIF
\STATE $\tau += (O_t, a_t, r_t, \text{done}_t, h_t)$
\ENDFOR
\STATE $\tau += (O_{T+1})$
\end{algorithmic}
\end{algorithm}

Below we discuss various components of Algorithm \ref{algo:rollout}.

\paragraph{Parallelism.}
Note that in Algorithm \ref{algo:MBRL}, we call Algorithm \ref{algo:rollout} in parallel, starting from $N_{\text{env}}$ observations $O^{1:N_{\text{env}}}_1$ (for environment rollout) or $\tilde{O}^{1:N_{\text{env}}}_1$ (for TWM rollout).

\paragraph{Burn-in.} The first time we collect data in the environment, we initialize the hidden state to zeros. 
The next time, we use burn-in to refresh the hidden state before rolling out the policy \citep{kapturowski2018recurrent}. 
We do so by passing the $M$ observations prior to $O_1$ to the policy, which updates the hidden state of the policy using the latest parameters.
(To use burn-in TWM rollout, we sample a trajectory of length $M+1$ in Step 4 of Algorithm \ref{algo:MBRL}.)
To enable burn-in, when collecting data, in Step 1 of Algorithm \ref{algo:MBRL}, we must also store the last $M$ environment observations $(O_{-M+1}, \ldots, O_0)$ prior to $O_1$.

\paragraph{TWM sampling.} As explained in the main text, sampling from the distribution $Q_{t+1} \sim p_{\Theta}(Q_{t+1} | Q_{1:t}, a_{1:t})$ is different when using (or not) block teacher forcing. For the former, the tokens of the next timestep $(q_{t+1}^1, \ldots, q_{t+1}^L)$ are sampled in parallel, while for the latter, they are sampled autoregressively.

\paragraph{Maximum buffer size.} To avoid running out of memory, we use a maximum buffer size and restrict the data buffer $\mathcal{D}$ in Algorithm \ref{algo:MBRL} to contain at most the last $128\text{k}$ observations. When the buffer is at capacity, we remove the oldest observations before adding the new ones. We use flashbax \citep{flashbax} to implement our replay buffer in JAX.

\subsubsection{World model update}

In practice, we decompose the world model updates into two steps. First, we update the tokenizer $N^{\text{iters}}_{\text{tok}}$ times. Second, we update the TWM $N^{\text{iters}}_{\text{TWM}}$ times. For both updates, we use $N_{\text{WM}}^{\text{mb training}}=3$ minibatches.
That is, Step 3 of Algorithm \ref{algo:MBRL} is implemented as in \cref{algo:step3}.

\begin{algorithm}[h!]
\small
\caption{Step 3 of Algorithm \ref{algo:MBRL}}
\begin{algorithmic}
\FOR{$\text{it}=1$ {\bfseries to} $N^{\text{iters}}_{\text{tok}}$}
\FOR{$k=1$ {\bfseries to} $N_{\text{WM}}^{\text{mb training}}$}
\STATE $N^{\text{start}} = (k-1) \left(N_{\text{env}} / N_{\text{WM}}^{\text{mb training}}\right) +1, ~~ N^{\text{end}} = k \left(N_{\text{env}} / N_{\text{WM}}^{\text{mb training}}\right) +1$
\STATE  $\tau_{\text{replay}} ^{n} =\text{sample-trajectory}(\mathcal{D}, T_{\text{WM}}), ~~~~~~~~~~~~~~~~~~~~~~~~~~~~~~~~~~~~~~~~~~~~~~ n=1:N_{\text{env}}$ \\
\STATE  $\Theta = \text{update-tokenizer}(\Theta, \tau_{\text{replay}} ^{N^{\text{start}}:N^{\text{end}}})$ with Equation \eqref{vqvae}\\
\ENDFOR
\ENDFOR
\FOR{$\text{it}=1$ {\bfseries to} $N^{\text{iters}}_{\text{TWM}}$}
\FOR{$k=1$ {\bfseries to} $N_{\text{WM}}^{\text{mb training}}$}
\STATE $N^{\text{start}} = (k-1) \left(N_{\text{env}} / N_{\text{WM}}^{\text{mb training}}\right) +1, ~~ N^{\text{end}} = k \left(N_{\text{env}} / N_{\text{WM}}^{\text{mb training}}\right) +1$
\STATE  $\tau_{\text{replay}} ^{n} =\text{sample-trajectory}(\mathcal{D}, T_{\text{WM}}), ~~~~~~~~~~~~~~~~~~~~~~~~~~~~~~~~~~~~~~~~~~~~~~ n=1:N_{\text{env}}$ \\
\STATE  $\Theta = \text{update-TWM}(\Theta, \tau_{\text{replay}} ^{N^{\text{start}}:N^{\text{end}}} )$ following Appendix \ref{sec:appendix_twm}\\
\ENDFOR
\ENDFOR
\end{algorithmic}
\label{algo:step3}
\end{algorithm}

We always set $N^{\text{iters}}_{\text{TWM}}=500$ to perform a large number of gradient updates. For M1-3, we set $N^{\text{iters}}_{\text{tok}}=500$ as well, but for M5 we reduce it to $N^{\text{iters}}_{\text{tok}}=25$ for the sake of speed---since NNT only adds new patches to the codebook.

\subsubsection{PPO policy update}
\label{sec:minibatch_epoch}

Finally, the PPO-policy-update procedure called in Steps 1 and 4 of Algorithm \ref{algo:MBRL} follows Algorithm \ref{algo:ppo_update}. 

When using PPO for MBRL, we found it critical to use different numbers of minibatches and different numbers of epochs on the trajectories collected on the environment and with TWM. 

In particular, as the trajectories collected in imagination are longer, we  reduce the number of parallel environments,
and use
$N_{\text{env}}^{\text{mb}}=8$ and $N_{\text{WM}}^{\text{mb}}=1$.
This guarantees that the PPO updates are on batches of comparable sizes---$6 \times 96$ for real trajectories, and $48 \times 20$ for imaginary trajectories.

In addition, while the environment trajectories are limited, we can simply rollout our TWM to collect more imaginary trajectories. Consequently, we set $N_{\text{env}}^{\text{epoch}}=4$, and $N_{\text{WM}}^{\text{epoch}}=1$. 

Finally, we do not use learning rate annealing for MBRL training.

\subsubsection{Hyperparameters}
Table \ref{table:mbfrl_high_level} summarizes the main parameters used in our MBRL training pipeline.
\begin{table*}[h!]
\small
\centering
\caption{MBRL main parameters.}
\label{table:mbfrl_high_level}
\begin{tabular}{p{0.55\textwidth}p{0.1\textwidth} }
\toprule
Hyperparameter & Value\\
\toprule
\toprule
Number of environments $N_{\text{env}}$ & $48$ \\
Rollout horizon in environment $T_{\text{env}}$ & $96$ \\
Rollout horizon for TWM $T_{\text{WM}}$ & $20$ \\
Burn-in horizon $M$ & $5$ \\
Buffer size & $128,000$ \\
Number of tokenizer updates $N^{\text{iters}}_{\text{tok}}$ (with VQ-VAE)  & $500$ \\
Number of tokenizer updates $N^{\text{iters}}_{\text{tok}}$ (with NNT) & $25$ \\
Number of TWM updates $N^{\text{iters}}_{\text{TWM}}$ & $500$ \\
Number of minibatches for TWM training $N_{\text{WM}}^{\text{mb training}}$ & $3$ \\
Background planning starting step $T_{\text{BP}}$ & $200\text{k}$ \\
Number of policy updates in imagination $N^{\text{iters}}_{\text{AC}}$ & $150$ \\
Number of PPO minibatches in environment $N_{\text{env}}^{\text{mb}}$ & $8$ \\
Number of PPO minibatches in imagination $N_{\text{WM}}^{\text{mb}}$ & $1$ \\
Number of epochs in environment $N_{\text{env}}^{\text{epoch}}$ & $4$ \\
Number of epochs in imagination $N_{\text{WM}}^{\text{epoch}}$ & $1$ \\
Learning rate annealing  & False \\
\bottomrule
\end{tabular}
\end{table*}

\newpage
\section{Comparing scores}

\label{sec:appendix_score}

\begin{figure*}[h!]
    \centering
    \begin{subfigure}[b]{0.48\textwidth}
        \centering
        \includegraphics[width=\textwidth]{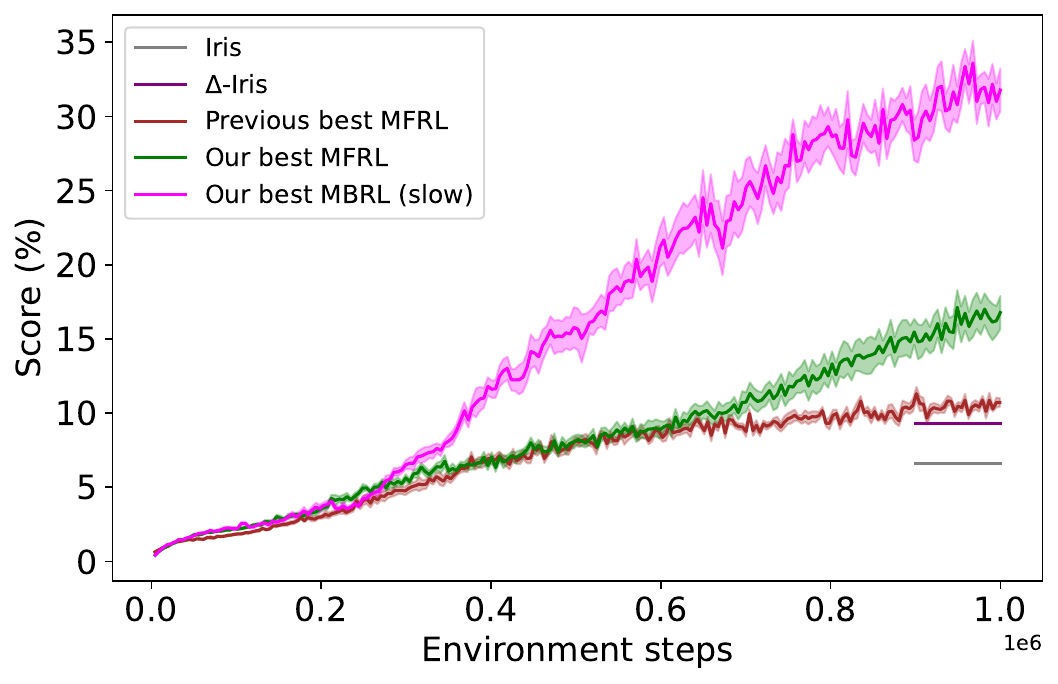}
    \end{subfigure}
    \hfill
    \begin{subfigure}[b]{0.48\textwidth}
        \centering
        \includegraphics[width=\textwidth]{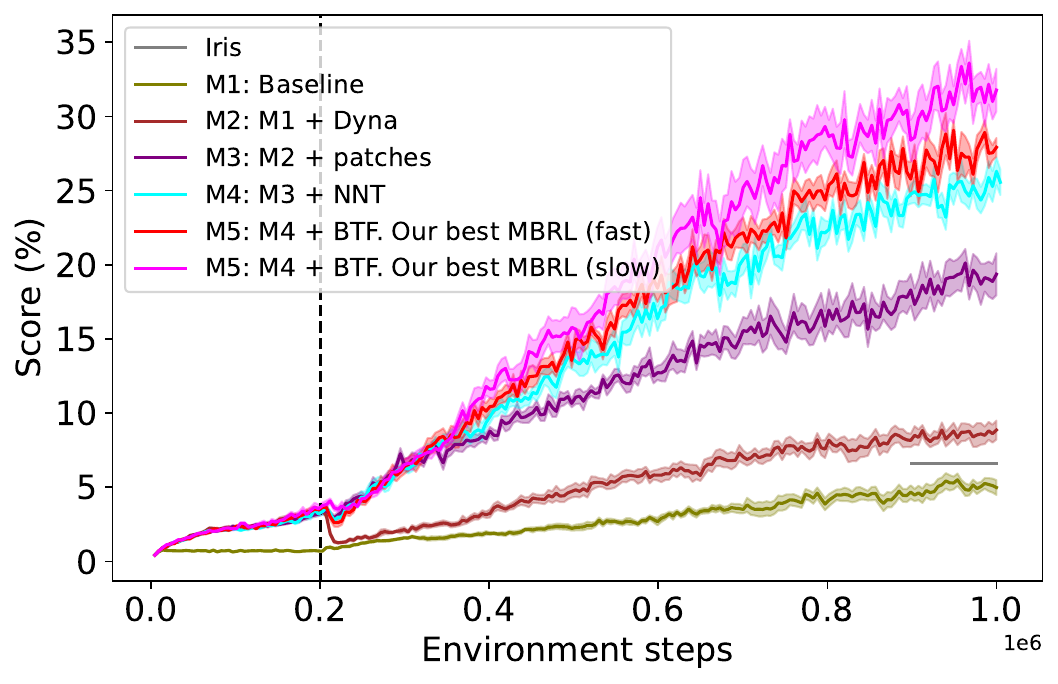}
    \end{subfigure}
    \caption{[Left] In addition to reaching higher rewards, our best MBRL and MFRL agents also achieve higher scores compared to the best previously published MBRL and MFRL results. [Right] MBRL agents' scores increase as they climb up the ladder of improvements.
    }
    \label{fig:score}
\vspace{-.75em}
\end{figure*}

Figure \ref{fig:score} completes the two main Figures \ref{fig:teaser}[left] and \ref{fig:main} by reporting the scores the different agents. Specifically, Figure \ref{fig:score}[left] compares our best MBRL and MFRL agents to the best previously published MBRL and MFRL agents. Figure \ref{fig:score}[right] displays the scores for the different agents on our ladder of improvements.

\bigskip

\bigskip
~
\bigskip

\section{Annealing the number of policy updates}
\label{sec:appendix_annealing}
\begin{figure}[h!]
\centering
\includegraphics[width=.5\linewidth]{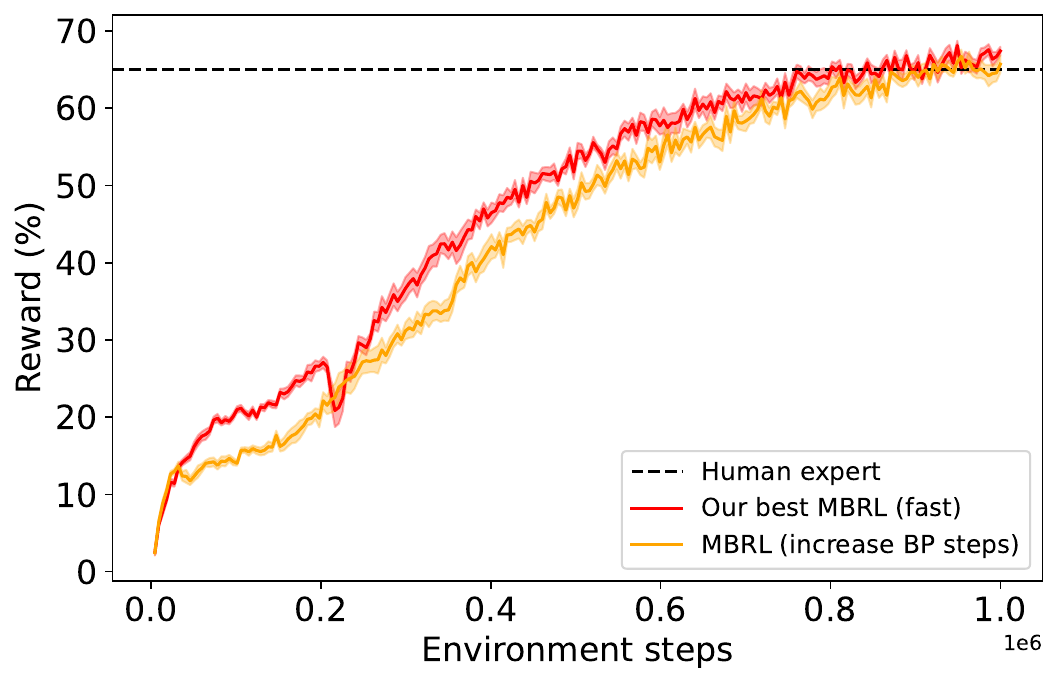}
\caption{Progressively increasing the number of policy updates in imagination from $N^{\text{iters}}_{\text{AC}}=0$ (when $T_{\text{total}}=0$ env. steps) to $N^{\text{iters}}_{\text{AC}}=300$ (when $T_{\text{total}}=1$M) removes the drop in performance observed otherwise when we start training in imagination.}
\label{fig:annealing}
\end{figure}

Figure \ref{fig:annealing} compares our best MFRL agent (with fast training) to an agent trained by annealing the number of policy updates in imaginary rollouts.

\newpage
\section{Additional world model comparisons}
\label{ap:symbol_accuracy}

This section complements Section \ref{sec:stationary} and presents two additional results to compare the different world models.

\begin{figure*}[h!]
    \centering
    \begin{tabular}{c}
        \includegraphics[width=0.4\textwidth]{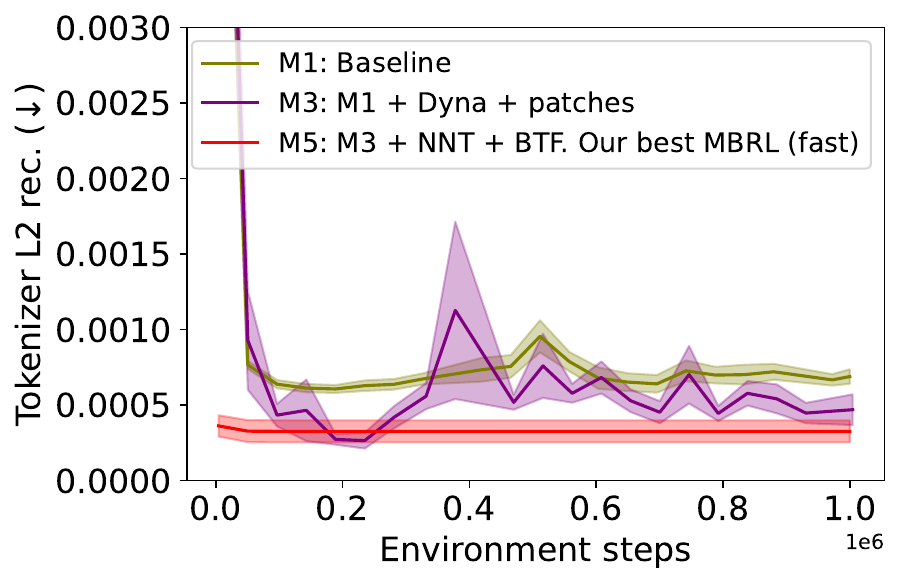}
    \end{tabular}
    \begin{tabular}{c}
        \includegraphics[width=0.4\textwidth]{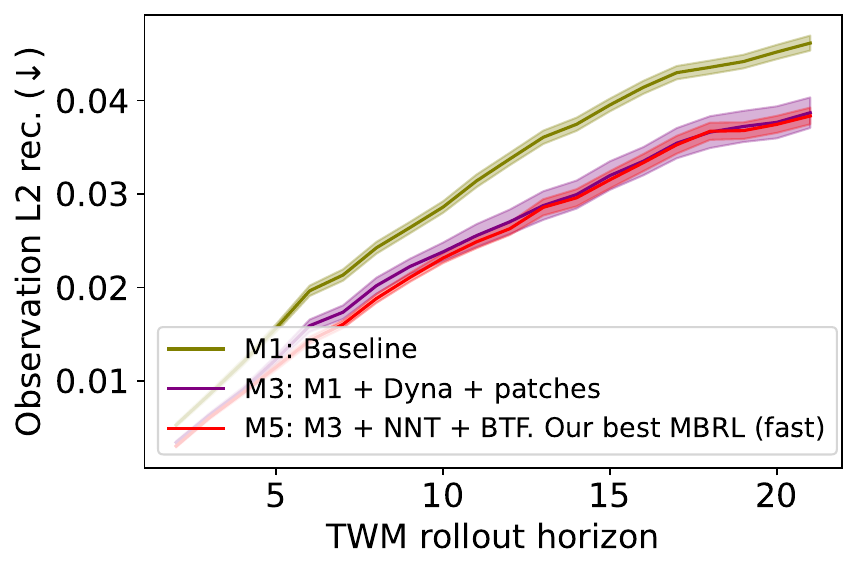}
    \end{tabular}
    \vspace{-.75em}
    \caption{TWM performance.[Left] Tokenizer L$2$ reconstruction error, averaged over rollouts. Lower is better. By construction, our best MBRL agent, which uses NNT, constantly reaches the lowest error, as NNT directly adds observation patches to its codebook. 
    [Right] TWM rollouts L$2$ observation reconstruction error, averaged over rollouts. Lower is better. M3 and M5, which both use patch factorization, achieve the lowest errors.
    }
    \label{fig:wm_quality_appendix}
\vspace{-.75em}
\end{figure*}

\subsection{Tokenizer reconstruction error}
We first use the evaluation dataset $\mathcal{D}_{\text{eval}}$ (introduced in Section \ref{sec:stationary}) to compare the tokenizer reconstruction error of our world models M1, M3, and M5---using the checkpoints at $1$M steps.
To do so, we independently encode and decode each observation $O^n_t \in \mathcal{D}_{\text{eval}}$, to obtain a tokenizer reconstruction $\hat{O}_t^{\text{tok},~n}$. Figure \ref{fig:wm_quality_appendix}[left] compares the average L$2$ reconstruction errors over the evaluation dataset:
$$
\frac{1}{(T + 1) N_{\text{eval}}} \sum_{n=1}^{N_{\text{eval}}} \sum_{t=1}^{T_{\text{eval}} +1} \| \hat{O}^{\text{tok},~n}_t - O^n_t \|_2^2,
$$
showing that all three models achieve low $L$2 reconstruction error. 
However our best model M5, which uses NNT, reaches a very low reconstruction error from the first iterations, since it directly adds image patches to its codebook rather than learning the codes online.

\subsection{Rollout reconstruction error}
Second, given a sequence of observations in a TWM rollout
$\hat{O}^{\text{TWM},~n}_{1:T_{\text{eval}} + 1}$, and the corresponding sequence of observations in the environment $O^{n}_{1:T_{\text{eval}} + 1}$ (which both have executed the same sequence of actions),
Figure \ref{fig:wm_quality_appendix}[right] compares the observation L$2$ reconstruction errors at each timestep $t$ (averaged over the evaluation dataset):
$$
\mathcal{E}_t = 
\frac{1}{N_{\text{eval}}} \sum_{n=1}^{N_{\text{eval}}} \| \hat{O}^{\text{TWM},~n}_t - O^n_t \|_2^2,
~~~\forall t.
$$
As expected, the errors increase with $t$ as mistakes compound over the rollout.
Our best method and M3, which both uses patch factorization, achieve the lowest reconstruction errors.

\eat{
\subsection{Symbol accuracy}
In this section, we leverage an appealing property of Craftax-classic: each observation $O_t$ comes with an array of ground truth symbols $S_t=(S_t^{1:R})$, with $R=145$. Given $100\text{k}$ pairs $(O_t, S_t)$, we learn a CNN $f_{\mu}$, to predict the symbols from the observation.
Next, we use $f_{\mu}$ to predict the symbols from the predicted rollout observation $\hat{O}^n_{1:T_{\text{eval}}+1}$. As above, we define the symbol accuracy at timestep $t$:
\begin{equation*}
    \mathcal{A}_t = \frac{1}{R . N_{\text{eval}}} \sum_{n=1}^{N_{\text{eval}}} \sum_{r=1}^{R}\mathbf{1}(f_{\mu}^r(\hat{O}^n_t), S^{r,n}_t), ~ \forall t
\end{equation*}
where $\mathbf{1}(x,y) = 1 ~\text{iff.}~ x=y $ (and $0$ o.w.), $S^{r,n}_t$ denotes the ground truth $r$th symbol from the array $S^{n}_t$ associated with $O^n_t$, and $f_{\mu}^r(\hat{O}^n_t)$ its prediction for the rollout $\hat{O}^n_t$.
Figure \ref{fig:wm_quality_appendix}[right] shows that, as expected, symbol accuracies decrease as the rollout horizon increases. It further highlights that our best model M5 is better able to capture the game dynamics.
}

\subsection{Symbol extractor architecture}
Herein, we discuss the symbol extractor architecture introduced in Section \ref{sec:stationary}. $f_{\mu}$ consists of (a) a first convolution layer with kernel size $7 \times 7$, stride of $7$, and channel size $128$, which extracts a feature for each patch, (b) a ReLU activation, (c) a second convolution layer with kernel size $1 \times 1$, a stride of $1$, and a channel size $128$, (d) a second ReLU activation, (e) a final linear layer which transforms the $3$D convolutional output into a $2$D array of logits of size $145 * 17=1345$---where $R=145$ is the number of ground truth symbols associated with each image of Craftax-classic and each symbol $S_t^r\in \{1, \ldots, 17\}$. The symbol extractor is trained with a cross-entropy loss between the predicted symbol logits and their ground truth values $S_t$, and achieves a $99.0\%$ validation accuracy.

\medskip

\subsection{Rollout comparison}
\label{ap:rollout_comparison}

\begin{figure}[h!]
    \centering
    \includegraphics[width=\linewidth]{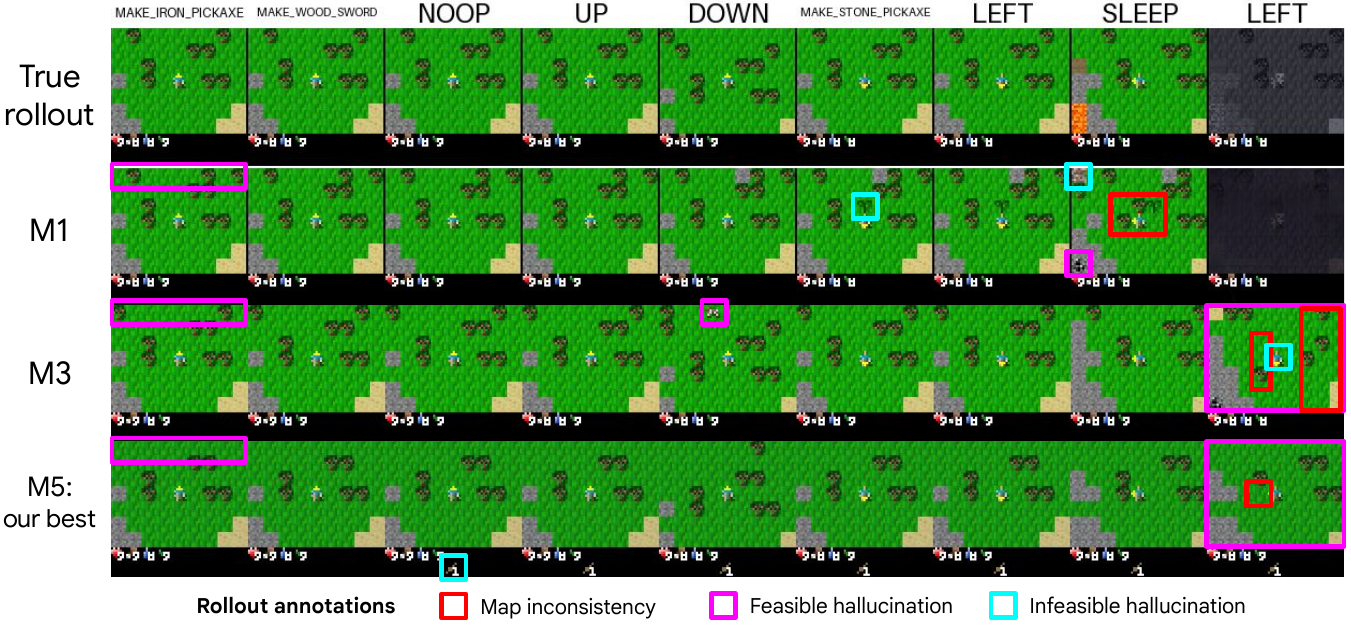}
    \caption{Additional rollout comparison for world models M1, M3 and M5. Best viewed zoomed in.
    \textbf{Map.}
    All models exhibit some map inconsistencies. M1 can make simple mistakes after the agent moves. Both M3 and M5 have map inconsistencies after the sleep actions, however the mistakes for M3 are far more severe.
    \textbf{Feasible hallucinations.}
    All models make feasible hallucinations when the agent exposes a new map region. 
    The sleep action is stochastic, and only sometimes results in the agent sleeping after taking the action. As a result, M3 and M5 make reasonable generations in predicting that the agent does not sleep in the final frame.
    \textbf{Infeasible hallucinations.}
    M1 generates cells that do not respect the game dynamics, such as spawning a plant without taking the place plant action, and creating a block type that cannot exist in that location.
    M3 turns the agent to face downwards without the down action. 
    M5 makes the wood sword despite the precondition of having wood inventory not being satisfied. 
    }
    \label{fig:more_rollouts}
\end{figure}

In Figure \ref{fig:more_rollouts}, we show an additional rollout that exhibits similar properties to those in Figure \ref{fig:rollout_eval}[right].
M1 and M3 make more simple mistakes in the map layout. 
All models generate predictions that can be inconsistent with the game dynamics. 
However the errors by M1 and M3 are more severe, as M5's mistake relates to the preconditions of the make sword action.

\newpage
\section{Comparing Craftax-classic and Craftax (full)}
\label{sec:classic_vs_full}
This section complements Section \ref{sec:craftax_full} and discusses the main differences between Craftax-classic and Craftax.
The first and second block Table \ref{tab:classic_vs_full} compares both environments. Note that we only use the first five parameters in our experiments in Section \ref{sec:craftax_full}.
The third and fourth blocks report the parameters used by our best MFRL and MBRL agents. In Craftax (full), for MFRL, we use $N_{\text{env}}=64$ environments and a rollout length $T_{\text{env}}=64$. 
Our SOTA MBRL agent uses $T_{\text{env}}=96$, $N_{\text{env}}=48$, and 
and $T_{\text{WM}}=20$ as in Craftax-classic. We reduced the buffer size to $48\text{k}$ to fit in GPU.
Our SOTA MBRL agent uses $T_{\text{env}}=96$ 
and $T_{\text{WM}}=20$ as in Craftax-classic, but reduces the number of environments to $N_{\text{env}}=16$ to fit in GPU.
All the others PPO parameters are the same as in Table \ref{table:mfrl_hyperparameters}.

\medskip

\begin{table*}[h!]
\small
    \centering
    \caption{Environment Craftax-classic vs Craftax (full).}
    \label{tab:classic_vs_full}
    \vspace{-.5em}
    \begin{tabular}{p{0.23\textwidth} p{0.35\textwidth}p{0.15\textwidth}p{0.15\textwidth}}
    \toprule
    Module & Hyperparameter & Classic & Full \\ 
    \toprule
    \toprule
    Environment (used)  & Image size & $63 \times 63$ & $130 \times 110$ \\
    & Patch size & $7 \times 7$ & $10 \times 10$ \\
    & Grid size & $9 \times 9$ & $13 \times 13$ \\
    & Action space size & $17$ & $43$ \\
    & Max reward (\# achievements) & $22$ & $226$ \\
    \midrule
    Environment (not used) & Symbolic (one-hot) input size & $1345$ & $8268$ \\
    & Max cardinality of each symbol & $17$ & $40$ \\
    & Number of levels & $1$ & $10$ \\
    \midrule
    \midrule
    MFRL parameters & Number of environments $N_{\text{env}}$ & $48$ & $64$ \\
    & Rollout horizon in environment $T_{\text{env}}$ & $96$ & $64$ \\
    \midrule
    MBRL parameters & Number of environments $N_{\text{env}}$ & $48$ & $16$ \\
    & Rollout horizon in environment $T_{\text{env}}$ & $96$ & $96$ \\
    & Rollout horizon for TWM $T_{\text{WM}}$ & $20$ & $20$ \\
    & Buffer size & $48,000$ & $128,000$ \\
    \bottomrule
    \end{tabular}
\end{table*}

\newpage
\section{Adapting Craftax-classic parameters to solve MinAtar}
\label{sec:appendix_minatar}

This section details the adaptations we made to our pipeline for solving the MinAtar environments, presented in Section \ref{sec:minatar}. First, Table \ref{tab:mfrl_minatar} outlines the modifications to our MFRL agent. Notably, we incorporate layer normalization \citep{ba2016layer} and Swish activation function \citep{ramachandran2017swish} within the ImpalaCNN architecture. Furthermore, we found it beneficial for the actor and critic networks to share weights up to their distinct final linear layers. We also adjust some PPO hyperparameters.

\begin{table*}[h!]
\centering
\caption{MFRL changes for MinAtar}
\label{tab:mfrl_minatar}
\vspace{-.5em}
\begin{tabular}{p{0.15\textwidth} p{0.35\textwidth}p{0.2\textwidth}p{0.2\textwidth}}
\toprule
Module & Parameter & Craftax & Minatar \\ 
\toprule
\toprule
Environment  & Image size & $63 \times 63 \times 3$ & $10\times10 \times K$ \\
\midrule
ImpalaCNN  & Normalization & Batch normalization & Layer normalization \\
& Activation & ReLU & Swish \\
& Shared network & False & True\\
\midrule
PPO & $\gamma$ & $0.925$ & $0.95$ \\
& $\lambda$ & $0.625$ & $0.75$ \\
& PPO target discount factor $\alpha$ & $0.95$ & $0.925$
\\
\bottomrule
\end{tabular}
\end{table*}

These modifications result in a solid MFRL agent, whose performance is detailed in Section \ref{sec:minatar}. We then develop our MBRL agent on top by implementing the changes outlined in Table \ref{tab:mbrl_minatar}. Specifically, we decompose each MinAtar image into $25$ patches of size $2 \times 2 \times K$ each. In addition, we increase (a) the number of TWM updates to from $500$ to $2$k and (b) the number of policy updates in imagination from $150$ to $2$k. Critically, to address the high cost of bad actions in certain games (e.g. Breakout), we assign a weight of $10$ to the cross-entropy losses of the reward and of the done states.
This strongly penalizes inaccurate predictions of terminal states in imaginary rollouts. Additionally, we observe a potential issue during training in imagination where the agent could collapse to output the same action consistently. To mitigate this ``action collapse'' and promote exploration, we increase the entropy coefficient in the imagination phase from $0.01$ to $0.05$.

\begin{table*}[h!]
\centering
\caption{MBRL changes for MinAtar}
\label{tab:mbrl_minatar}
\vspace{-.5em}
\begin{tabular}{p{0.15\textwidth} p{0.35\textwidth}p{0.15\textwidth}p{0.15\textwidth}}
\toprule
Module & Parameter & Craftax & Minatar \\ 
\toprule
\toprule
Tokenizer & Patch size & $7 \times 7 \times 3$ & $2 \times 2 \times K$ \\
& Grid size & $9 \times 9$ & $5 \times 5$ \\
\midrule
Training & Number of policy updates $N^{\text{iters}}_{\text{AC}}$ & $150$ & $2,000$ \\
& Number of TWM updates $N^{\text{iters}}_{\text{TWM}}$ & $500$ & $2,000$ \\
& Termination and reward weight & $1$ & $10$ \\
& PPO entropy coeff. in imagination  & $0.01$ & $0.05$\\
\bottomrule
\end{tabular}
\end{table*}

\medskip 

Note that all the MinAtar games use the same hyperparameters.

\newpage
\section{Transformer World Models for multiplayer OpenSpiel games}\label{sec:appendix_twm_multiplayer}

\paragraph{Game characteristics:} Table \ref{tab:openspiel_env_params} details each OpenSpiel game characteristics.
Observations are represented as categorical 1D arrays. 

\begin{table*}[h!]
\centering
\caption{OpenSpiel environment parameters.}
\label{tab:openspiel_env_params}
\vspace{-.25em}
\begin{tabular}{p{0.3\textwidth} p{0.15\textwidth}p{0.15\textwidth}p{0.15\textwidth}}
\toprule
 & Tic-tac-toe & Leduc Poker & Bargaining \\ 
\toprule
\toprule
Observation size & $27$ & $16$ & $93$ \\
Number of categories & $2$ & $14$ & $2$ \\
Number of player actions & $9$ & $3$ & $121$ \\
Number of chance actions & $0$ & $6$ & $1002$ \\
Partially observed & False & True & True \\
Reward range & $\{-1, 0, 1\}$ & $\{-1, 0, 1\}$ & $[0, 10]$\\
\bottomrule
\end{tabular}
\end{table*}

\paragraph{TWM details:}
As detailed in Section \ref{sec:openspiel}, our goal is to train a single agent (either Player 1 or Player 2) under the assumption that its opponent uniformly picks a legal action. While extending our MFRL pipeline is straightforward, our MBRL pipeline requires additional work to guarantee that our TWM generates rollouts that respect the game rules and correctly model both players' actions

To achieve this, when deploying our policy in the environment, we collect at each timestep (a) the current player ID ($0$ for the chance player, $1$ for Player 1, $2$ for Player 2), (b) both players' observations, (c) the set of legal actions for the current player (including their probabilities for the chance player), (d) the action taken by the current player.

The TWM is now given sequences of the form $\{x_t^1, x_t^2, a_t\}_{1\le t \le T}$, where $x_t^1$ (resp. $x_t^2$) is the observation of Player 1 (resp. Player 2), and $a_t$ can be an action of Player 1, of Player 2, or of the chance player. We use distinct actions range for each player: Player 1's actions $a_t \in \{1, \ldots N^{\text{player actions}} \}$, Player 2's actions $a_t \in \{ N^{\text{player actions}} + 1, \ldots , 2 * N^{\text{player actions}} \}$ and the chance player's actions $a_t \in \{ 2 N^{\text{player actions}}, \ldots, 2 N^{\text{player actions}} + N^{\text{chance actions}} \}$. This design guarantees that the current player ID can be inferred from the current action.

TWM is trained with six cross-entropy losses, expanding our presentation in Appendix \ref{sec:appendix_twm}:

\smallskip
$\textbf{(1)}$
The cross-entropy for the reward prediction. We use a one-hot encoding of the reward target, with $3$ categories for Tic-Tac-Toe and Leduc Poker, and $11$ categories for Bargaining.

\smallskip
$\textbf{(2)}$
The cross-entropy for the termination predictions. 

\smallskip
$\textbf{(3}$ 
The cross-entropy for the next symbol predictions, where the symbols vary between $1$ and the number of categories for each game.

\smallskip
$\textbf{(4)}$ 
The cross-entropy for the next player ID predictions.

\smallskip
$\textbf{(5}$ 
The cross-entropy for the next set of (binary) legal actions. We individually predict the logit of each action being a legal one for the next move.

\smallskip
$\textbf{(6}$ 
The cross-entropy for the next chance actions.

This training approach guarantees that TWM can generate imaginary rollouts of the form $\{\hat{x}_t^1, \hat{x}_t^2, \hat{a}_t\}_{1\le t \le T}$ which respect the game rules and accurately model the actions of both players and of the chance player. During policy training, we disregard the other agent's predicted observation sequence. For instance, a Player 1 policy is trained on $\{\hat{x}_t^1, \hat{a}_t\}_{1\le t \le T}$.

\paragraph{Hyperparameters:}
Table \ref{tab:mbrl_openspiel} details the hyperparameters used for OpenSpiel games. The MFRL parameters are mostly similar to Craftax. We train until both players have, in total, taken $T_{\text{total}}=100\text{k}$ actions. Note that for Tic-Tac-Toe, which is fully visible, we drop the RNN and use an MLP policy. Similarly, the MBRL parameters are largely inherited from Minatar environments. In particular, both the number of TWM updates and the number of policy updates in imagination are set to $2$k. We do not use warmup ($T_{\text{BP}}=0$) before starting training the policy in imagination.

\begin{table*}[h!]
\centering
\caption{Parameters for OpenSpiel}
\label{tab:mbrl_openspiel}
\vspace{-.5em}
\begin{tabular}{p{0.15\textwidth} p{0.4\textwidth}p{0.25\textwidth}}
\toprule
Module & Parameter & OpenSpiel \\ 
\toprule
\toprule
ImpalaCNN  & Normalization & Batch normalization \\
& Activation & ReLU \\
& Shared network & True\\
& Use RNN & iff. partially observed \\
\midrule
PPO & $\gamma$ & $0.925$ \\
& $\lambda$ & $0.625$ \\
& PPO target discount factor $\alpha$ & $0.95$ \\
\midrule
Training & Total number of steps $T_{\text{total}}$ & $100\text{k}$ \\
& Background planning starting step $T_{\text{BP}}$ & $0$ \\
& Number of policy updates $N^{\text{iters}}_{\text{AC}}$ & $2,000$ \\
& Number of TWM updates $N^{\text{iters}}_{\text{TWM}}$ & $2,000$ \\
& Termination and reward weight & $1$ \\
& PPO entropy coeff. in imagination  & $0.05$\\
\bottomrule
\end{tabular}
\end{table*}

%% file: main.bbl
\begin{thebibliography}{61}
\providecommand{\natexlab}[1]{#1}
\providecommand{\url}[1]{\texttt{#1}}
\expandafter\ifx\csname urlstyle\endcsname\relax
  \providecommand{\doi}[1]{doi: #1}\else
  \providecommand{\doi}{doi: \begingroup \urlstyle{rm}\Url}\fi

\bibitem[Agarwal et~al.(2024)Agarwal, Andrews, and Kahou]{agarwal2024learning}
P.~Agarwal, S.~Andrews, and S.~E. Kahou.
\newblock Learning to play atari in a world of tokens.
\newblock \emph{ICML}, 2024.

\bibitem[Alonso et~al.(2024)Alonso, Jelley, Micheli, Kanervisto, Storkey,
  Pearce, and Fleuret]{alonso2024diffusion}
E.~Alonso, A.~Jelley, V.~Micheli, A.~Kanervisto, A.~Storkey, T.~Pearce, and
  F.~Fleuret.
\newblock Diffusion for world modeling: Visual details matter in atari.
\newblock In \emph{The Thirty-eighth Annual Conference on Neural Information
  Processing Systems}, 2024.
\newblock URL \url{https://openreview.net/forum?id=NadTwTODgC}.

\bibitem[Alver and Precup(2024)]{Alver2024}
S.~Alver and D.~Precup.
\newblock A look at value-based decision-time vs. background planning methods
  across different settings.
\newblock In \emph{Seventeenth European Workshop on Reinforcement Learning},
  Oct. 2024.
\newblock URL \url{https://openreview.net/pdf?id=Vx2ETvHId8}.

\bibitem[Ba et~al.(2016)Ba, Kiros, and Hinton]{ba2016layer}
J.~L. Ba, J.~R. Kiros, and G.~E. Hinton.
\newblock Layer normalization.
\newblock \emph{arXiv preprint arXiv:1607.06450}, 2016.

\bibitem[Bradbury et~al.(2018)Bradbury, Frostig, Hawkins, Johnson, Leary,
  Maclaurin, Necula, Paszke, Vander{P}las, Wanderman-{M}ilne, and
  Zhang]{jax2018github}
J.~Bradbury, R.~Frostig, P.~Hawkins, M.~J. Johnson, C.~Leary, D.~Maclaurin,
  G.~Necula, A.~Paszke, J.~Vander{P}las, S.~Wanderman-{M}ilne, and Q.~Zhang.
\newblock {JAX}: composable transformations of {P}ython+{N}um{P}y programs,
  2018.
\newblock URL \url{http://github.com/jax-ml/jax}.

\bibitem[Chen et~al.(2022)Chen, Wu, Yoon, and Ahn]{chen2202transdreamer}
C.~Chen, Y.-F. Wu, J.~Yoon, and S.~Ahn.
\newblock Transdreamer: Reinforcement learning with transformer world models.
\newblock \emph{URL http://arxiv. org/abs/2202}, 9481, 2022.

\bibitem[Cohen et~al.(2024)Cohen, Wang, Kang, and Mannor]{cohen2024improving}
L.~Cohen, K.~Wang, B.~Kang, and S.~Mannor.
\newblock Improving token-based world models with parallel observation
  prediction.
\newblock \emph{arXiv preprint arXiv:2402.05643}, 2024.

\bibitem[Cohen et~al.(2025)Cohen, Wang, Kang, Gadot, and Mannor]{cohen2025text}
L.~Cohen, K.~Wang, B.~Kang, U.~Gadot, and S.~Mannor.
\newblock M3: A modular world model over streams of tokens.
\newblock \emph{arXiv preprint arXiv:2502.11537}, 2025.

\bibitem[Espeholt et~al.(2018{\natexlab{a}})Espeholt, Soyer, Munos, Simonyan,
  Mnih, Ward, Doron, Firoiu, Harley, Dunning, Legg, and
  Kavukcuoglu]{Espeholt2018}
L.~Espeholt, H.~Soyer, R.~Munos, K.~Simonyan, V.~Mnih, T.~Ward, Y.~Doron,
  V.~Firoiu, T.~Harley, I.~Dunning, S.~Legg, and K.~Kavukcuoglu.
\newblock {IMPALA}: Scalable distributed deep-{RL} with importance weighted
  actor-learner architectures.
\newblock In \emph{ICML}, pages 1407--1416. PMLR, July 2018{\natexlab{a}}.
\newblock URL \url{https://proceedings.mlr.press/v80/espeholt18a.html}.

\bibitem[Espeholt et~al.(2018{\natexlab{b}})Espeholt, Soyer, Munos, Simonyan,
  Mnih, Ward, Doron, Firoiu, Harley, Dunning, et~al.]{espeholt2018impala}
L.~Espeholt, H.~Soyer, R.~Munos, K.~Simonyan, V.~Mnih, T.~Ward, Y.~Doron,
  V.~Firoiu, T.~Harley, I.~Dunning, et~al.
\newblock Impala: Scalable distributed deep-rl with importance weighted
  actor-learner architectures.
\newblock In \emph{International conference on machine learning}, pages
  1407--1416. PMLR, 2018{\natexlab{b}}.

\bibitem[Farebrother et~al.(2024)Farebrother, Orbay, Vuong, Taiga, Chebotar,
  Xiao, Irpan, Levine, Castro, Faust, Kumar, and Agarwal]{Farebrother2024}
J.~Farebrother, J.~Orbay, Q.~Vuong, A.~A. Taiga, Y.~Chebotar, T.~Xiao,
  A.~Irpan, S.~Levine, P.~S. Castro, A.~Faust, A.~Kumar, and R.~Agarwal.
\newblock Stop regressing: Training value functions via classification for
  scalable deep {RL}.
\newblock In \emph{Forty-first International Conference on Machine Learning},
  June 2024.
\newblock URL \url{https://openreview.net/pdf?id=dVpFKfqF3R}.

\bibitem[Guan et~al.(2024)Guan, Verch, Voelcker, Jackson, Papernot, and
  Cunningham]{guan2024temporal}
J.~Guan, S.~Verch, C.~Voelcker, E.~Jackson, N.~Papernot, and W.~Cunningham.
\newblock Temporal-difference learning using distributed error signals.
\newblock \emph{Advances in Neural Information Processing Systems},
  37:\penalty0 108710--108734, 2024.

\bibitem[Ha and Schmidhuber(2018{\natexlab{a}})]{Ha2018}
D.~Ha and J.~Schmidhuber.
\newblock World models.
\newblock In \emph{NIPS}, 2018{\natexlab{a}}.
\newblock URL \url{http://arxiv.org/abs/1803.10122}.

\bibitem[Ha and Schmidhuber(2018{\natexlab{b}})]{ha2018recurrent}
D.~Ha and J.~Schmidhuber.
\newblock Recurrent world models facilitate policy evolution.
\newblock \emph{Advances in neural information processing systems}, 31,
  2018{\natexlab{b}}.

\bibitem[Hafner(2021)]{hafner2021benchmarking}
D.~Hafner.
\newblock Benchmarking the spectrum of agent capabilities.
\newblock \emph{arXiv preprint arXiv:2109.06780}, 2021.

\bibitem[Hafner et~al.(2020{\natexlab{a}})Hafner, Lillicrap, Ba, and
  Norouzi]{Hafner2020}
D.~Hafner, T.~Lillicrap, J.~Ba, and M.~Norouzi.
\newblock Dream to control: Learning behaviors by latent imagination.
\newblock In \emph{ICLR}, 2020{\natexlab{a}}.
\newblock URL \url{https://openreview.net/forum?id=S1lOTC4tDS}.

\bibitem[Hafner et~al.(2020{\natexlab{b}})Hafner, Lillicrap, Norouzi, and
  Ba]{hafner2020mastering}
D.~Hafner, T.~Lillicrap, M.~Norouzi, and J.~Ba.
\newblock Mastering atari with discrete world models.
\newblock \emph{arXiv preprint arXiv:2010.02193}, 2020{\natexlab{b}}.

\bibitem[Hafner et~al.(2023)Hafner, Pasukonis, Ba, and
  Lillicrap]{hafner2023mastering}
D.~Hafner, J.~Pasukonis, J.~Ba, and T.~Lillicrap.
\newblock Mastering diverse domains through world models.
\newblock \emph{arXiv preprint arXiv:2301.04104}, 2023.

\bibitem[Hansen et~al.(2024)Hansen, Su, and Wang]{Hansen2024}
N.~Hansen, H.~Su, and X.~Wang.
\newblock {TD}-{MPC2}: Scalable, robust world models for continuous control.
\newblock 2024.
\newblock URL \url{http://arxiv.org/abs/2310.16828}.

\bibitem[He et~al.(2016)He, Zhang, Ren, and Sun]{he2016deep}
K.~He, X.~Zhang, S.~Ren, and J.~Sun.
\newblock Deep residual learning for image recognition.
\newblock In \emph{Proceedings of the IEEE conference on computer vision and
  pattern recognition}, pages 770--778, 2016.

\bibitem[Hessel et~al.(2018)Hessel, Modayil, van Hasselt, Schaul, Ostrovski,
  Dabney, Horgan, Piot, Azar, and Silver]{Hessel2018}
M.~Hessel, J.~Modayil, H.~van Hasselt, T.~Schaul, G.~Ostrovski, W.~Dabney,
  D.~Horgan, B.~Piot, M.~Azar, and D.~Silver.
\newblock Rainbow: Combining improvements in deep reinforcement learning.
\newblock In \emph{AAAI}, 2018.
\newblock URL \url{http://arxiv.org/abs/1710.02298}.

\bibitem[Holland et~al.(2018)Holland, Talvitie, and Bowling]{Holland2018}
G.~Z. Holland, E.~J. Talvitie, and M.~Bowling.
\newblock The effect of planning shape on dyna-style planning in
  high-dimensional state spaces.
\newblock \emph{arXiv [cs.AI]}, June 2018.
\newblock URL \url{http://arxiv.org/abs/1806.01825}.

\bibitem[Ioffe and Szegedy(2015)]{ioffe2015batch}
S.~Ioffe and C.~Szegedy.
\newblock Batch normalization: Accelerating deep network training by reducing
  internal covariate shift.
\newblock In \emph{International conference on machine learning}, pages
  448--456. pmlr, 2015.

\bibitem[Kaiser et~al.(2019)Kaiser, Babaeizadeh, Milos, Osinski, Campbell,
  Czechowski, Erhan, Finn, Kozakowski, Levine, Mohiuddin, Sepassi, Tucker, and
  Michalewski]{Kaiser2019}
L.~Kaiser, M.~Babaeizadeh, P.~Milos, B.~Osinski, R.~H. Campbell, K.~Czechowski,
  D.~Erhan, C.~Finn, P.~Kozakowski, S.~Levine, A.~Mohiuddin, R.~Sepassi,
  G.~Tucker, and H.~Michalewski.
\newblock Model-based reinforcement learning for atari.
\newblock \emph{arXiv [cs.LG]}, Mar. 2019.
\newblock URL \url{http://arxiv.org/abs/1903.00374}.

\bibitem[Kapturowski et~al.(2018)Kapturowski, Ostrovski, Quan, Munos, and
  Dabney]{kapturowski2018recurrent}
S.~Kapturowski, G.~Ostrovski, J.~Quan, R.~Munos, and W.~Dabney.
\newblock Recurrent experience replay in distributed reinforcement learning.
\newblock In \emph{International conference on learning representations}, 2018.

\bibitem[Kauvar et~al.(2023)Kauvar, Doyle, Zhou, and Haber]{Kauvar2023}
I.~Kauvar, C.~Doyle, L.~Zhou, and N.~Haber.
\newblock Curious replay for model-based adaptation.
\newblock In \emph{ICML}, June 2023.
\newblock URL \url{https://arxiv.org/abs/2306.15934}.

\bibitem[Kingma(2014)]{kingma2014adam}
D.~P. Kingma.
\newblock Adam: A method for stochastic optimization.
\newblock \emph{arXiv preprint arXiv:1412.6980}, 2014.

\bibitem[Lambert et~al.(2022)Lambert, Pister, and Calandra]{Lambert2022}
N.~Lambert, K.~Pister, and R.~Calandra.
\newblock Investigating compounding prediction errors in learned dynamics
  models.
\newblock \emph{arXiv [cs.LG]}, Mar. 2022.
\newblock URL \url{http://arxiv.org/abs/2203.09637}.

\bibitem[Lanctot et~al.(2019)Lanctot, Lockhart, Lespiau, Zambaldi, Upadhyay,
  P\'{e}rolat, Srinivasan, Timbers, Tuyls, Omidshafiei, Hennes, Morrill,
  Muller, Ewalds, Faulkner, Kram\'{a}r, Vylder, Saeta, Bradbury, Ding,
  Borgeaud, Lai, Schrittwieser, Anthony, Hughes, Danihelka, and
  Ryan-Davis]{LanctotEtAl2019OpenSpiel}
M.~Lanctot, E.~Lockhart, J.-B. Lespiau, V.~Zambaldi, S.~Upadhyay,
  J.~P\'{e}rolat, S.~Srinivasan, F.~Timbers, K.~Tuyls, S.~Omidshafiei,
  D.~Hennes, D.~Morrill, P.~Muller, T.~Ewalds, R.~Faulkner, J.~Kram\'{a}r,
  B.~D. Vylder, B.~Saeta, J.~Bradbury, D.~Ding, S.~Borgeaud, M.~Lai,
  J.~Schrittwieser, T.~Anthony, E.~Hughes, I.~Danihelka, and J.~Ryan-Davis.
\newblock {OpenSpiel}: A framework for reinforcement learning in games.
\newblock \emph{CoRR}, abs/1908.09453, 2019.
\newblock URL \url{http://arxiv.org/abs/1908.09453}.

\bibitem[Lei~Ba et~al.(2016)Lei~Ba, Kiros, and Hinton]{lei2016layer}
J.~Lei~Ba, J.~R. Kiros, and G.~E. Hinton.
\newblock Layer normalization.
\newblock \emph{ArXiv e-prints}, pages arXiv--1607, 2016.

\bibitem[Lewis et~al.(2017)Lewis, Yarats, Dauphin, Parikh, and
  Batra]{lewis2017deal}
M.~Lewis, D.~Yarats, Y.~N. Dauphin, D.~Parikh, and D.~Batra.
\newblock Deal or no deal? end-to-end learning for negotiation dialogues.
\newblock \emph{arXiv preprint arXiv:1706.05125}, 2017.

\bibitem[Lu et~al.(2022)Lu, Kuba, Letcher, Metz, Schroeder~de Witt, and
  Foerster]{lu2022discovered}
C.~Lu, J.~Kuba, A.~Letcher, L.~Metz, C.~Schroeder~de Witt, and J.~Foerster.
\newblock Discovered policy optimisation.
\newblock \emph{Advances in Neural Information Processing Systems},
  35:\penalty0 16455--16468, 2022.

\bibitem[Matthews et~al.(2024)Matthews, Beukman, Ellis, Samvelyan, Jackson,
  Coward, and Foerster]{matthews2024craftax}
M.~Matthews, M.~Beukman, B.~Ellis, M.~Samvelyan, M.~Jackson, S.~Coward, and
  J.~Foerster.
\newblock Craftax: A lightning-fast benchmark for open-ended reinforcement
  learning.
\newblock \emph{arXiv preprint arXiv:2402.16801}, 2024.

\bibitem[Micheli et~al.(2022)Micheli, Alonso, and
  Fleuret]{micheli2022transformers}
V.~Micheli, E.~Alonso, and F.~Fleuret.
\newblock Transformers are sample-efficient world models.
\newblock \emph{arXiv preprint arXiv:2209.00588}, 2022.

\bibitem[Micheli et~al.(2024)Micheli, Alonso, and
  Fleuret]{micheli2024efficient}
V.~Micheli, E.~Alonso, and F.~Fleuret.
\newblock Efficient world models with context-aware tokenization.
\newblock \emph{arXiv preprint arXiv:2406.19320}, 2024.

\bibitem[Mirzasoleiman et~al.(2020)Mirzasoleiman, Bilmes, and
  Leskovec]{Mirzasoleiman2020}
B.~Mirzasoleiman, J.~Bilmes, and J.~Leskovec.
\newblock Coresets for data-efficient training of machine learning models.
\newblock In \emph{ICML}, 2020.
\newblock URL
  \url{http://proceedings.mlr.press/v119/mirzasoleiman20a/mirzasoleiman20a.pdf}.

\bibitem[Moerland et~al.(2023)Moerland, Broekens, Plaat, and
  Jonker]{Moerland2023}
T.~M. Moerland, J.~Broekens, A.~Plaat, and C.~M. Jonker.
\newblock Model-based reinforcement learning: A survey.
\newblock \emph{Foundations and Trends in Machine Learning}, 16\penalty0
  (1):\penalty0 1--118, 2023.
\newblock URL \url{https://arxiv.org/abs/2006.16712}.

\bibitem[Moon et~al.(2024)Moon, Yeom, Park, and Song]{moon2024discovering}
S.~Moon, J.~Yeom, B.~Park, and H.~O. Song.
\newblock Discovering hierarchical achievements in reinforcement learning via
  contrastive learning.
\newblock \emph{Advances in Neural Information Processing Systems}, 36, 2024.

\bibitem[Murphy(2024)]{murphy2024reinforcement}
K.~Murphy.
\newblock Reinforcement learning: An overview.
\newblock \emph{arXiv preprint arXiv:2412.05265}, 2024.

\bibitem[Ni et~al.(2024)Ni, Eysenbach, Seyedsalehi, Ma, Gehring, Mahajan, and
  Bacon]{Ni2024}
T.~Ni, B.~Eysenbach, E.~Seyedsalehi, M.~Ma, C.~Gehring, A.~Mahajan, and P.-L.
  Bacon.
\newblock Bridging state and history representations: Understanding
  self-predictive {RL}.
\newblock In \emph{ICLR}, Jan. 2024.
\newblock URL \url{http://arxiv.org/abs/2401.08898}.

\bibitem[OpenDILab()]{awesome_mbrl}
OpenDILab.
\newblock {Awesome Model-Based Reinforcement Learning}.
\newblock \url{https://github.com/opendilab/awesome-model-based-RL}.

\bibitem[Oquab et~al.(2024)Oquab, Darcet, Moutakanni, Vo, Szafraniec, Khalidov,
  Fernandez, Haziza, Massa, El-Nouby, Assran, Ballas, Galuba, Howes, Huang, Li,
  Misra, Rabbat, Sharma, Synnaeve, Xu, Jegou, Mairal, Labatut, Joulin, and
  Bojanowski]{Oquab2024}
M.~Oquab, T.~Darcet, T.~Moutakanni, H.~V. Vo, M.~Szafraniec, V.~Khalidov,
  P.~Fernandez, D.~Haziza, F.~Massa, A.~El-Nouby, M.~Assran, N.~Ballas,
  W.~Galuba, R.~Howes, P.-Y. Huang, S.-W. Li, I.~Misra, M.~Rabbat, V.~Sharma,
  G.~Synnaeve, H.~Xu, H.~Jegou, J.~Mairal, P.~Labatut, A.~Joulin, and
  P.~Bojanowski.
\newblock {DINOv2}: Learning robust visual features without supervision.
\newblock \emph{Transactions on Machine Learning Research}, 2024.
\newblock URL \url{https://openreview.net/forum?id=a68SUt6zFt}.

\bibitem[Radford et~al.(2019)Radford, Wu, Child, Luan, Amodei, Sutskever,
  et~al.]{radford2019language}
A.~Radford, J.~Wu, R.~Child, D.~Luan, D.~Amodei, I.~Sutskever, et~al.
\newblock Language models are unsupervised multitask learners.
\newblock \emph{OpenAI blog}, 1\penalty0 (8):\penalty0 9, 2019.

\bibitem[Ramachandran et~al.(2017)Ramachandran, Zoph, and
  Le]{ramachandran2017swish}
P.~Ramachandran, B.~Zoph, and Q.~V. Le.
\newblock Swish: a self-gated activation function.
\newblock \emph{arXiv preprint arXiv:1710.05941}, 7\penalty0 (1):\penalty0 5,
  2017.

\bibitem[Ravi et~al.(2024)Ravi, Gabeur, Hu, Hu, Ryali, Ma, Khedr, R{\"a}dle,
  Rolland, Gustafson, et~al.]{ravi2024sam}
N.~Ravi, V.~Gabeur, Y.-T. Hu, R.~Hu, C.~Ryali, T.~Ma, H.~Khedr, R.~R{\"a}dle,
  C.~Rolland, L.~Gustafson, et~al.
\newblock Sam 2: Segment anything in images and videos.
\newblock \emph{arXiv preprint arXiv:2408.00714}, 2024.

\bibitem[Robine et~al.(2023)Robine, H{\"o}ftmann, Uelwer, and
  Harmeling]{robine2023transformer}
J.~Robine, M.~H{\"o}ftmann, T.~Uelwer, and S.~Harmeling.
\newblock Transformer-based world models are happy with 100k interactions.
\newblock \emph{arXiv preprint arXiv:2303.07109}, 2023.

\bibitem[Schrittwieser et~al.(2020)Schrittwieser, Antonoglou, Hubert, Simonyan,
  Sifre, Schmitt, Guez, Lockhart, Hassabis, Graepel,
  et~al.]{schrittwieser2020mastering}
J.~Schrittwieser, I.~Antonoglou, T.~Hubert, K.~Simonyan, L.~Sifre, S.~Schmitt,
  A.~Guez, E.~Lockhart, D.~Hassabis, T.~Graepel, et~al.
\newblock Mastering atari, go, chess and shogi by planning with a learned
  model.
\newblock \emph{Nature}, 588\penalty0 (7839):\penalty0 604--609, 2020.

\bibitem[Schulman et~al.(2015)Schulman, Moritz, Levine, Jordan, and
  Abbeel]{schulman2015high}
J.~Schulman, P.~Moritz, S.~Levine, M.~Jordan, and P.~Abbeel.
\newblock High-dimensional continuous control using generalized advantage
  estimation.
\newblock \emph{arXiv preprint arXiv:1506.02438}, 2015.

\bibitem[Schulman et~al.(2017)Schulman, Wolski, Dhariwal, Radford, and
  Klimov]{schulman2017proximal}
J.~Schulman, F.~Wolski, P.~Dhariwal, A.~Radford, and O.~Klimov.
\newblock Proximal policy optimization algorithms.
\newblock \emph{arXiv preprint arXiv:1707.06347}, 2017.

\bibitem[Schwarzer et~al.(2023)Schwarzer, Obando-Ceron, Courville, Bellemare,
  Agarwal, and Castro]{Schwarzer2023}
M.~Schwarzer, J.~Obando-Ceron, A.~Courville, M.~Bellemare, R.~Agarwal, and
  P.~S. Castro.
\newblock Bigger, better, faster: Human-level atari with human-level
  efficiency.
\newblock In \emph{ICML}, May 2023.
\newblock URL \url{http://arxiv.org/abs/2305.19452}.

\bibitem[Southey et~al.(2012)Southey, Bowling, Larson, Piccione, Burch,
  Billings, and Rayner]{southey2012bayes}
F.~Southey, M.~P. Bowling, B.~Larson, C.~Piccione, N.~Burch, D.~Billings, and
  C.~Rayner.
\newblock Bayes' bluff: Opponent modelling in poker.
\newblock \emph{arXiv preprint arXiv:1207.1411}, 2012.

\bibitem[Su et~al.(2024)Su, Ahmed, Lu, Pan, Bo, and Liu]{su2024roformer}
J.~Su, M.~Ahmed, Y.~Lu, S.~Pan, W.~Bo, and Y.~Liu.
\newblock Roformer: Enhanced transformer with rotary position embedding.
\newblock \emph{Neurocomputing}, 568:\penalty0 127063, 2024.

\bibitem[Sun et~al.(2023)Sun, Dong, Huang, Ma, Xia, Xue, Wang, and
  Wei]{sun2023retentive}
Y.~Sun, L.~Dong, S.~Huang, S.~Ma, Y.~Xia, J.~Xue, J.~Wang, and F.~Wei.
\newblock Retentive network: A successor to transformer for large language
  models.
\newblock \emph{arXiv preprint arXiv:2307.08621}, 2023.

\bibitem[Sutton(1990)]{sutton1990integrated}
R.~S. Sutton.
\newblock Integrated architectures for learning, planning, and reacting based
  on approximating dynamic programming.
\newblock In \emph{Machine learning proceedings 1990}, pages 216--224.
  Elsevier, 1990.

\bibitem[Sutton and Barto(2018)]{sutton2018reinforcement}
R.~S. Sutton and A.~G. Barto.
\newblock \emph{Reinforcement learning: An introduction}.
\newblock MIT press, 2018.

\bibitem[Toledo et~al.(2023)Toledo, Midgley, Byrne, Tilbury, Macfarlane,
  Courtot, and Laterre]{flashbax}
E.~Toledo, L.~Midgley, D.~Byrne, C.~R. Tilbury, M.~Macfarlane, C.~Courtot, and
  A.~Laterre.
\newblock Flashbax: Streamlining experience replay buffers for reinforcement
  learning with jax, 2023.
\newblock URL \url{https://github.com/instadeepai/flashbax/}.

\bibitem[Van Den~Oord et~al.(2017)Van Den~Oord, Vinyals, et~al.]{van2017neural}
A.~Van Den~Oord, O.~Vinyals, et~al.
\newblock Neural discrete representation learning.
\newblock \emph{Advances in neural information processing systems}, 30, 2017.

\bibitem[Van~Hasselt et~al.(2019)Van~Hasselt, Hessel, and
  Aslanides]{van2019use}
H.~P. Van~Hasselt, M.~Hessel, and J.~Aslanides.
\newblock When to use parametric models in reinforcement learning?
\newblock \emph{Advances in Neural Information Processing Systems}, 32, 2019.

\bibitem[Ye et~al.(2021)Ye, Liu, Kurutach, Abbeel, and Gao]{Ye2021}
W.~Ye, S.~Liu, T.~Kurutach, P.~Abbeel, and Y.~Gao.
\newblock Mastering atari games with limited data.
\newblock In \emph{NIPS}, Nov. 2021.
\newblock URL \url{https://openreview.net/pdf?id=OKrNPg3xR3T}.

\bibitem[{Young} and {Tian}(2019)]{young19minatar}
K.~{Young} and T.~{Tian}.
\newblock Minatar: An atari-inspired testbed for thorough and reproducible
  reinforcement learning experiments.
\newblock \emph{arXiv preprint arXiv:1903.03176}, 2019.

\bibitem[Zhang et~al.(2024)Zhang, Wang, Sun, Yuan, and Huang]{zhang2024storm}
W.~Zhang, G.~Wang, J.~Sun, Y.~Yuan, and G.~Huang.
\newblock Storm: Efficient stochastic transformer based world models for
  reinforcement learning.
\newblock \emph{Advances in Neural Information Processing Systems}, 36, 2024.

\end{thebibliography}
